\documentclass[pdflatex,sn-mathphys-num]{sn-jnl}%

\usepackage{graphicx}%
\usepackage{multirow}%
\usepackage{amsmath,amssymb,amsfonts}%
\usepackage{amsthm}%
\usepackage{mathrsfs}%
\usepackage[title]{appendix}%
\usepackage{xcolor}%
\usepackage{textcomp}%
\usepackage{manyfoot}%
\usepackage{booktabs}%
\usepackage{algorithm}%
\usepackage{algorithmicx}%
\usepackage{algpseudocode}%
\usepackage{listings}%
\usepackage{graphicx}
\usepackage{subcaption}
\usepackage{longtable}
\usepackage{adjustbox}  %
\usepackage{tabularx}   %
\usepackage{float}
\usepackage{placeins}

\usepackage[utf8]{inputenc}
\usepackage[T1]{fontenc}
\usepackage{lineno}
\usepackage{textcomp}
\usepackage{tabularx}
\usepackage{array}
\usepackage{ragged2e}

\DeclareUnicodeCharacter{0252}{a}  %
\DeclareUnicodeCharacter{0279}{r}  %
\DeclareUnicodeCharacter{0254}{o}  %

\usepackage[most]{tcolorbox}
\usepackage{listings}

\newtcolorbox{promptbox}[1]{
    colback=gray!3,       %
    colframe=gray!80,     %
    arc=1mm,              %
    boxrule=0.5pt,        %
    leftrule=3pt,         %
    fonttitle=\small\bfseries\sffamily,
    coltitle=black,       %
    attach title to upper,
    after title={\par\smallskip},
    title=#1,
    breakable
}

\theoremstyle{thmstyleone}%

\theoremstyle{thmstyletwo}%

\theoremstyle{thmstylethree}%

\begin{document}
\title[Multi-agent AI creativity]{Multi-agent AI systems outperform human teams in creativity}

\author[1]{\fnm{Tiancheng} \sur{Hu}}\email{th656@cam.ac.uk}

\author[2,3]{\fnm{Yixuan} \sur{Jiang}}\email{jiangyixuan0813@gmail.com}

\author[4]{\fnm{Haotian} \sur{Li}}\email{haotian.li@microsoft.com}

\author[5,6]{\fnm{José} \sur{Hernández-Orallo}}\email{jh2135@cam.ac.uk}

\author[4]{\fnm{Xing} \sur{Xie}}\email{xing.xie@microsoft.com}

\author[1]{\fnm{Nigel} \sur{Collier}}\email{nhc30@cam.ac.uk}

\author*[3]{\fnm{David} \sur{Stillwell}}\email{d.stillwell@jbs.cam.ac.uk}

\author*[3]{\fnm{Luning} \sur{Sun}}\email{ls523@cam.ac.uk}

\affil[1]{\orgdiv{Department of Theoretical and Applied Linguistics}, \orgname{University of Cambridge}, \city{Cambridge}, \postcode{ CB3 9DA}, \country{United Kingdom}}

\affil[2]{\orgdiv{Department of Psychology and Behavioral Sciences}, \orgname{Zhejiang University}, \city{Hangzhou}, \postcode{310058}, \country{China}}

\affil*[3]{\orgdiv{The Psychometrics Centre, Cambridge Judge Business School}, \orgname{University of Cambridge},  \city{Cambridge}, \postcode{CB2 1AG},  \country{United Kingdom}}

\affil[4]{ \orgname{Microsoft Research Asia},  \city{Beijing}, \postcode{100080},  \country{China}}

\affil[5]{\orgdiv{Leverhulme Centre for the Future of Intelligence}, \orgname{University of Cambridge},  \city{Cambridge}, \postcode{CB2 1SB},  \country{United Kingdom}}

\affil[6]{\orgdiv{Valencian Research Institute for Artificial Intelligence (VRAIN)}, \orgname{Universitat Politècnica de València},  \city{València}, \postcode{46022},  \country{Spain}}


\maketitle

\begin{abstract}

Although artificial intelligence (AI) now matches or exceeds human performance across numerous cognitive tasks, creativity remains a highly contested frontier. As AI systems based on large language models (LLMs) are increasingly adopted in research and innovation, it is essential to understand and augment their creativity. Here we demonstrate that multi-agent LLM teams not only surpass single agents, but also substantially outperform human teams in creativity (Cohen's $d=1.50$) across 4,541 multi-agent LLM ideas and 341 human-team ideas on six diverse problem-solving tasks. This advantage is driven by novelty while maintaining comparable usefulness. To investigate the generative processes in both groups, we represent conversations as paths through semantic space using neural language model representations. Both LLM and human teams produce more creative ideas when conversations range widely rather than staying centered on a single theme (low global coherence). However, the additional patterns that predict creativity differ: LLM teams benefit from efficient exploration (high semantic spread, shorter paths), while human teams benefit from maintaining smooth conversational flow (high local coherence, frequent pivots). Additionally, we identify model choice and discussion structure as orthogonal design levers that together explain 26.8\% of variance in LLM conversational dynamics, paving the way for systematic approaches to developing multi-agent systems with augmented creative capabilities.

\end{abstract}

\section*{Introduction}

Creativity—the capacity to generate ideas that are both novel and useful—has long been considered a defining feature of human intelligence~\cite{amabile1996creativity,boden2004creative, runco2012standard}. For decades, creativity remained the domain where human superiority over machines seemed assured, often described as a hallmark of human intelligence and a “final frontier” for artificial intelligence (AI) ~\cite{boden2004creative,colton2012computational}, even as AI systems surpassed structured, closed-ended tasks like image classification, medical diagnostics, and standardized academic exams~\cite{DBLP:conf/iccv/HeZRS15, esteva2017dermatologist, openai2024gpt4technicalreport}. Yet as large language models (LLMs) increasingly participate in knowledge work—from software engineering to scientific research~\cite{peng2023impactaideveloperproductivity,paradis2024doesaiimpactdevelopment,novikov2025alphaevolvecodingagentscientific,Asai2026,lu2026towards}, understanding whether and how artificial agents can be genuinely creative becomes a major scientific question.

Multi-agent systems have already demonstrated transformative potential across AI domains. From AlphaGo's self-play training \cite{Silver2017} to multi-agent reinforcement learning in complex games \cite{openai2019dota2largescale}, interactions among artificial agents have produced capabilities beyond those achievable by isolated agents. This pattern echoes findings from human studies: groups can exhibit measurable collective intelligence that is not reducible to the ability of any single member \cite{woolley2010evidence}. Multi-agent LLM systems, where multiple AI instances discuss, critique and refine ideas \cite{10.5555/3692070.3692537, liang-etal-2024-encouraging, qian-etal-2024-chatdev,lin-etal-2025-creativity}, extend this paradigm to artificial agents. Yet whether this approach can generate more creative outcomes than single agents and human teams remains unknown. %

Here we present a systematic comparison of creativity in human teams and multi-agent LLM systems, evaluating 4,541 ideas from LLMs and 341 ideas from human teams across 71 experimental conditions and six diverse creative problem-solving tasks. All ideas were blindly evaluated by human judges following the standard definition of creativity \cite{runco2012standard} using the Consensual Assessment Technique \cite{amabile1982social}. To understand the generative processes underlying creative outcomes, we represent each conversational turn as a vector using neural language model representations, then characterize the resulting trajectory—how broadly it explores semantic space, how coherently it progresses and how it converges toward solutions.

We find that multi-agent LLM teams substantially outperform human teams in creativity (Cohen's $d=1.50$), driven by novelty advantages while maintaining comparable usefulness. Trajectory analysis reveals different patterns: both LLM and human teams produce more creative ideas when conversations range widely, but LLM teams additionally benefit from efficient exploration, while human teams additionally benefit from maintaining smooth conversational flow. Critically, model choice and discussion structure function as complementary design levers, together explaining 26.8\% of variance in LLM conversational dynamics. This demonstrates that the exploration strategies driving LLMs' creative advantage can be optimized, providing actionable mechanisms for designing multi-agent ideation systems.

\section*{LLM Teams Substantially Outperform Humans in Creativity}
\subsection*{LLM Teams Generate More Creative Outputs than Single Agents and Human Teams}
We compared creative outputs from human teams ($N=341$ ideas from 114 teams of three or six members) and multi-agent LLM teams ($N=4{,}541$ ideas from teams of three or six agents) across six creative problem-solving tasks varying in domain and realism (see Methods). We also collected 179 single-agent LLM ideas as a reference baseline. All ideas were mixed and evaluated blind to source by five human raters on novelty and usefulness, with overall creativity operationalized as their product, consistent with standard definitions in creativity research \citep{runco2012standard}.

\begin{figure}[t]
    \centering

    \begin{subfigure}[b]{0.95\textwidth}
        \centering
        \includegraphics[width=\linewidth]{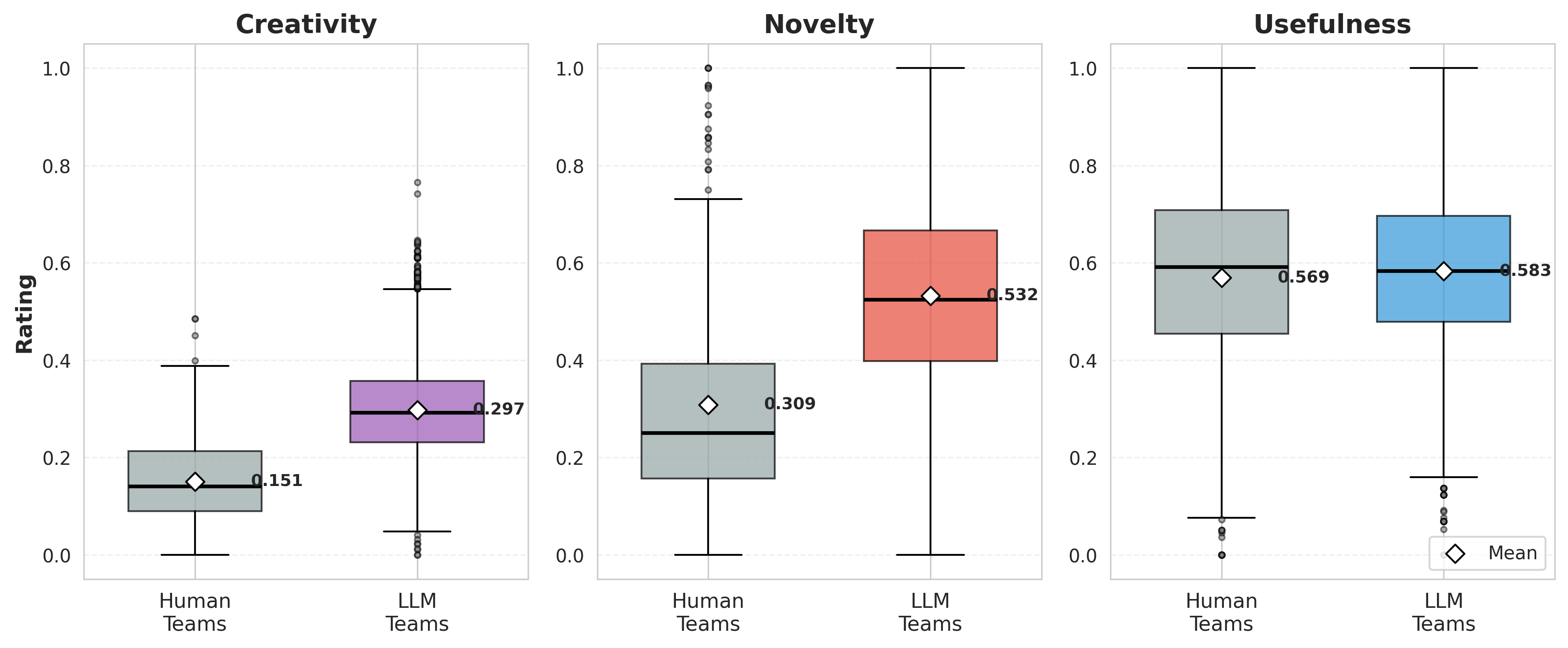} 
        \caption{Ratings by dimension}
        \label{fig:performance_a}
    \end{subfigure}
    
    \vspace{0.5em}
    
    \begin{subfigure}[b]{0.32\textwidth}
        \centering
        \includegraphics[width=\linewidth]{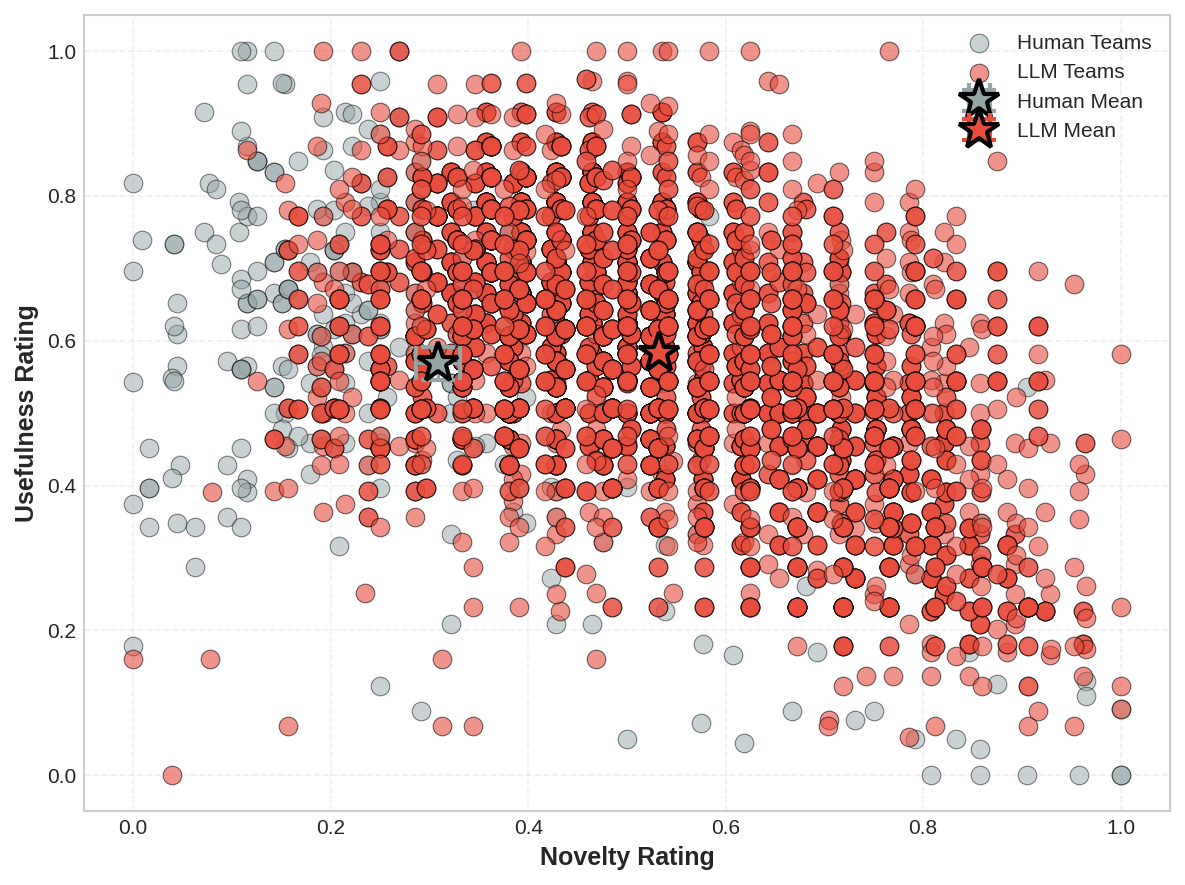}
        \caption{Novelty-usefulness trade-off}
        \label{fig:performance_b}
    \end{subfigure}
    \hfill
    \begin{subfigure}[b]{0.32\textwidth}
        \centering
        \includegraphics[width=\linewidth]{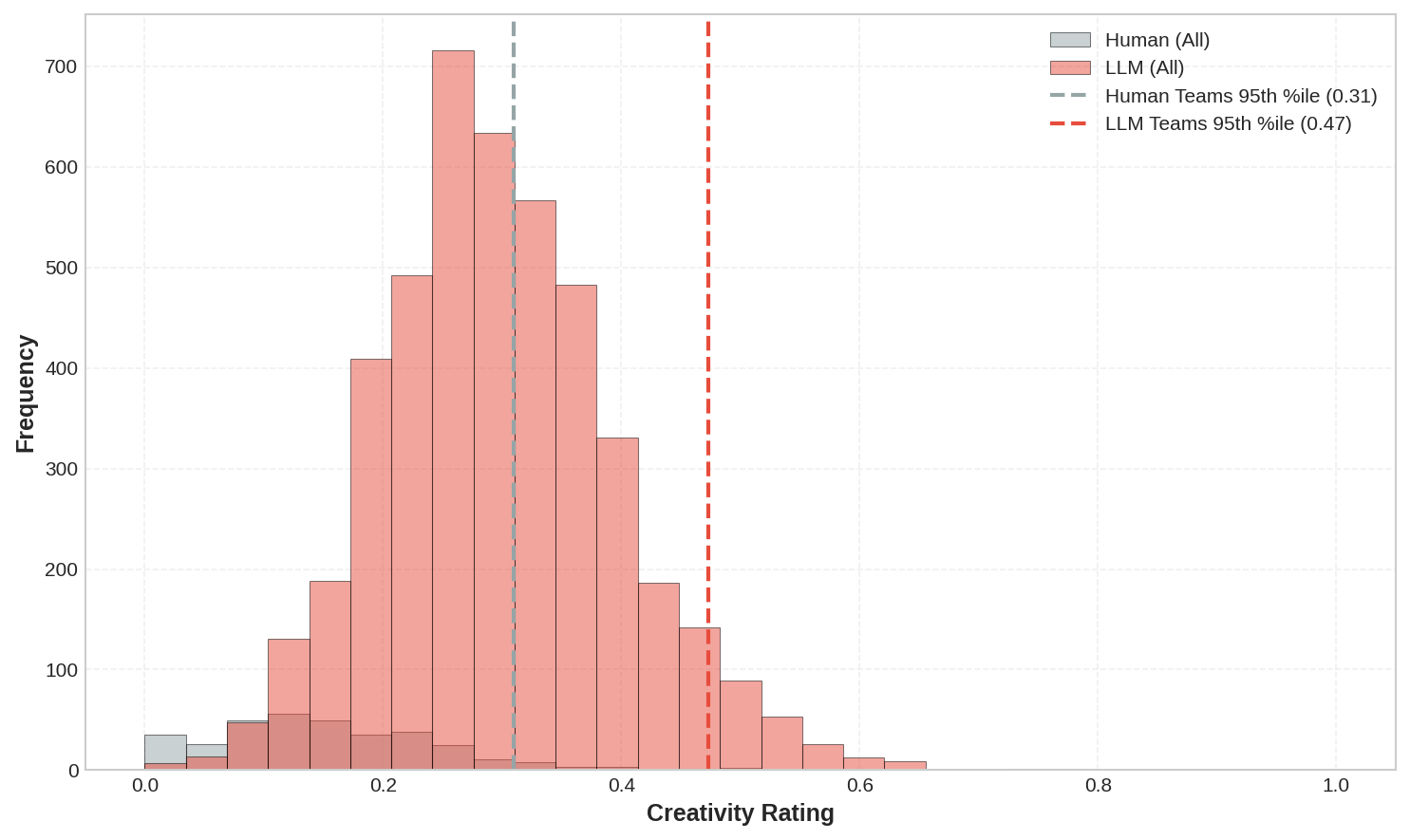}
        \caption{Distribution with 95th percentile}
        \label{fig:performance_c}
    \end{subfigure}
    \hfill
    \begin{subfigure}[b]{0.32\textwidth}
        \centering
        \includegraphics[width=\linewidth]{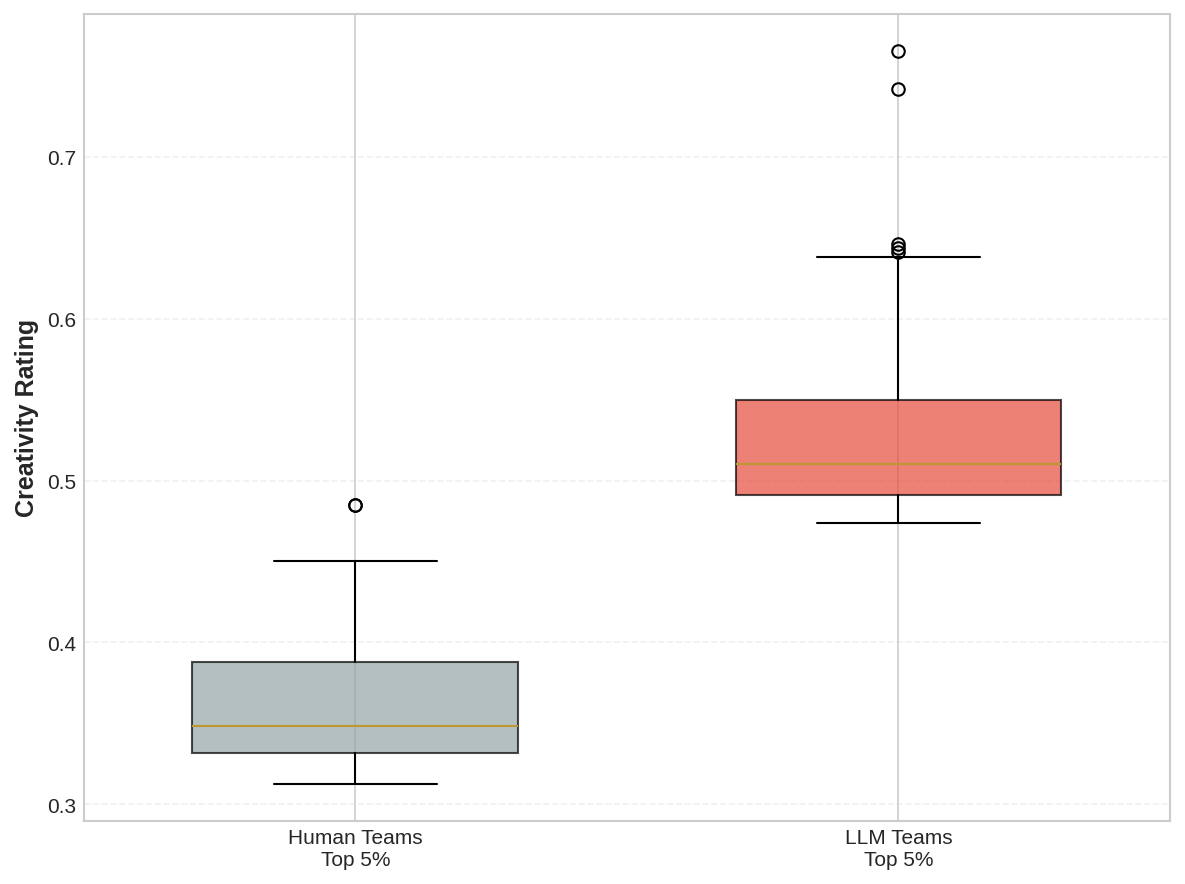}
        \caption{Top 5\% comparison}
        \label{fig:performance_d}
    \end{subfigure}
    
    \caption{\textbf{LLM teams substantially outperform human teams across multiple creativity dimensions.} 
    (\textbf{A}) Distribution of creativity, novelty, and usefulness ratings for human teams and multi-agent LLM teams. Box plots show median (line), interquartile range (box), and 1.5 times the interquartile range (whiskers); diamonds indicate means. LLM teams show large advantages in creativity ($d=1.50$) and novelty ($d=1.29$) while maintaining comparable usefulness ($d=0.08$, $p=0.142$). 
    (\textbf{B}) Distribution of ideas in novelty-usefulness space. Each point represents one idea, colored by source (gray = human, red = LLM). Stars indicate group means with 95\% confidence intervals.
    (\textbf{C}) Distribution of all ideas with 95th percentile thresholds marked (dashed lines). The LLM distribution is shifted rightward relative to the human distribution. 
    (\textbf{D}) Box plots comparing the top 5\% of ideas from each group (LLM: $M=0.53$; Human: $M=0.37$).}
    \label{fig:performance}
\end{figure}

Aggregating across models and discussion structures, LLM teams substantially outperformed human teams in overall creativity (LLM: $M=0.297$, 95\% CI [0.295, 0.300]; Human: $M=0.151$, 95\% CI [0.141, 0.160];  two-sided independent-samples $t(4880)=-26.73$, $p<0.001$, Cohen's $d=1.50$; Figure~\ref{fig:performance_a}). This comparison remained significant in task-adjusted, clustered, and mixed-effects robustness models accounting for the nested structure of ideas within tasks and generating systems (see Section~\ref{sec:robustness}). This advantage was also robust to alternative creativity operationalization (additive: $d=1.37$; see Section~\ref{sec:robustness}). This large effect was driven primarily by a marked advantage in novelty: LLM ideas were rated substantially more novel than human ideas (LLM: $M=0.532$, 95\% CI [0.527, 0.537]; Human: $M=0.309$, 95\% CI [0.286, 0.331]; $t(4880)=-22.97$, $p<0.001$, $d=1.29$). This novelty advantage came without sacrificing usefulness: LLM and human ideas were rated comparably on this dimension (LLM: $M=0.583$, 95\% CI [0.578, 0.587]; Human: $M=0.569$, 95\% CI [0.546, 0.592]; $t(4880)=-1.47$, $p=0.14$, $d=0.08$).

To contextualize multi-agent performance relative to individual model capability, we compared each model's single-agent output ($N=30$ per model) against its best-performing multi-agent configuration. While single LLM agents substantially outperformed human teams (single-agent $M=0.253$--$0.306$ vs.\ human $M=0.151$), the best multi-agent configuration significantly improved upon these already-high baselines for all three models tested in both settings: GPT-4.1 ($M=0.298$ vs.\ $0.253$, $d=0.52$, $p=0.006$), o3-low ($M=0.345$ vs.\ $0.289$, $d=0.61$, $p=0.003$), and o3-high ($M=0.345$ vs.\ $0.306$, $d=0.39$, $p=0.04$).

The six tasks spanned societal challenges (plastic waste, educational inequality), business problems (supply chain, employee attrition), and deliberately hypothetical scenarios (sorry pandemic, singing in shower), the latter serving to mitigate potential memorization concerns. LLM teams consistently outperformed human teams across all tasks (Table~\ref{tab:descriptive_stats_overall}), demonstrating that this advantage generalizes across domains. A novelty-usefulness trade-off analysis (Figure~\ref{fig:performance_b}) revealed that LLM teams occupy a distinct region of the creativity space: they can generate ideas that are both highly novel and practically useful (upper right quadrant), whereas human ideas exhibit a salient trade-off pattern—ideas tend to be either novel or useful, but rarely both.

Beyond mean differences, the LLM advantage extended to the upper tail of the creativity distribution, a critical consideration given that real-world applications often seek the single best solution rather than average performance. Figure~\ref{fig:performance_c} shows that the entire LLM distribution was shifted rightward relative to the human distribution, with a substantially higher 95th percentile threshold (LLM: 0.47; Human: 0.31, a 53\% increase). The top-scoring LLM idea (creativity = 0.77) exceeded the top human idea (0.48) by 58\%. Even among top-rated ideas, LLM teams maintained their advantage (Figure~\ref{fig:performance_d}): the top 5\% of LLM ideas ($M=0.53$, $SD=0.05$) substantially exceeded the top 5\% of human ideas ($M=0.37$, $SD=0.06$). This indicated that the LLM advantage extended beyond average performance to the upper tail of the evaluated idea distribution.

\subsection*{Discussion Structure Shapes Model Creativity}

Discussion structure provided a way to test how interaction shapes creative output, beyond simply comparing single agents with multiple agents. Different protocols varied in how much they encourage idea pooling, critique, revision, and convergence, allowing us to ask which forms of interaction are most associated with higher-quality creative solutions.

\begin{figure}[ht]
\centering
\includegraphics[width=0.85\textwidth]{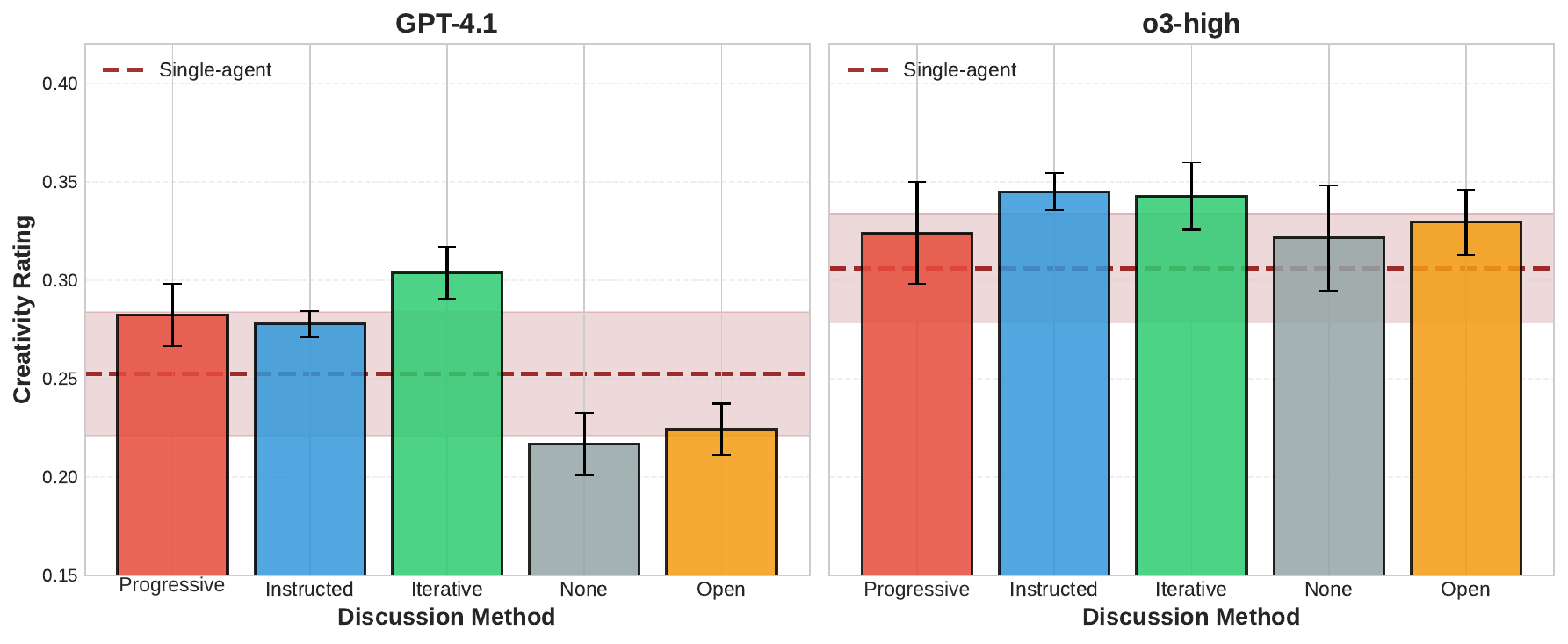}
\caption{\textbf{Discussion structure effects vary by model type.} Creativity ratings across discussion methods. The dashed line
  and shaded region indicate the single-agent mean and 95\% confidence interval (CI), respectively; error bars denote 95\% CIs for
  multi-agent means. For GPT-4.1 (left), iterative refinement significantly outperforms the None (no discussion) condition ($M =
  0.308$ vs.\ $M = 0.212$, $\beta = +0.097$, $p < 0.001$). Conversely, o3-high (right) maintains consistently high creativity across
  all structures ($M = 0.322$--$0.345$), with only instructed discussion showing a small task-adjusted improvement over None ($\beta
  = +0.023$, $p = 0.047$; other structured conditions $p > 0.05$).}

\label{fig:discussion_model_interaction}
\end{figure}

We tested five discussion structures: Open Discussion (unstructured brainstorming with a summary at the end), Instructed Discussion (structured discussion cycles with options to agree, modify, or replace ideas), Iterative Refinement (proposal-and-selection cycles proposing a single new idea and comparing it to existing ideas), Progressive Improvement (explicit separation of divergent generation prioritizing novelty and convergent refinement enhancing usefulness), and None (no discussion following idea generation and rating). The resulting creative performance varied substantially by model type (Figure~\ref{fig:discussion_model_interaction}).

For GPT-4.1, discussion structure significantly affected team creativity. Compared to no discussion ($M=0.212$), iterative refinement produced the largest gain ($M=0.308$, $\beta=+0.097$, $p<0.001$), followed by progressive ($\beta=+0.084$, $p<0.001$) and instructed discussions ($\beta=+0.065$, $p<0.001$). In contrast, o3-high showed minimal sensitivity to discussion structure, with all protocols producing similar levels of creativity ($M=0.322$--$0.345$, most $p>0.05$). The model $\times$ discussion interaction was highly significant ($p<0.001$), suggesting that reasoning models may already perform internal deliberation that reduces the need for external discussion scaffolding. Consistent with this interpretation, sensitivity to discussion structure decreased monotonically with reasoning effort: creativity varied by 0.096 across discussion structures for GPT-4.1, 0.038 for o3-low, and only 0.023 for o3-high (Table~\ref{tab:descriptive_stats_discussion_methods}), indicating a dose-response relationship between internal reasoning capacity and dependence on external scaffolding.

Decomposing creativity into novelty and usefulness revealed a trade-off: discussion structures that boosted novelty systematically reduced usefulness, and vice versa. GPT-4.1's iterative discussion partially overcame this trade-off by substantially boosting novelty ($\beta=+0.208$, $p<0.001$) without proportionally reducing usefulness (see Table~\ref{tab:descriptive_stats_discussion_methods} for full decomposition). These discussion structure effects remained significant after controlling for total token count ($\Delta R^2 = 0.06\%$, token $\beta \approx 0$, n.s.), confirming that creativity gains reflect discussion dynamics rather than inference-time scaling (see Table~\ref{tab:robustness_tokens}).

\subsection*{Agent Personas Shift Novelty-Usefulness Balance Without Improving Creativity}

Comparing GPT-4.1 teams with same, different, or no persona assignments in matched conditions, we found no significant difference in overall creativity (one-way ANOVA: $F=2.09$, $p=0.124$). Different-persona teams performed comparably to no-persona teams ($M=0.267$ vs.\ $0.269$; two-sided $t(958)=0.25$, $p=0.803$), whereas same-persona teams showed a small, marginally lower creativity score than no-persona teams ($M=0.258$ vs.\ $0.269$; two-sided $t(956)=1.91$, $p=0.057$). Persona assignment nevertheless shifted the novelty-usefulness balance: relative to no-persona teams (Novelty: $M=0.541$; Usefulness: $M=0.534$), same-persona teams produced less novel but more useful ideas (Novelty: $M=0.461$, $t(956)=7.15$, $p<0.001$; Usefulness: $M=0.587$, $t(956)=-4.98$, $p<0.001$), and different-persona teams show the same pattern (Novelty: $M=0.480$, $t(958)=5.31$, $p<0.001$; Usefulness: $M=0.585$, $t(958)=-4.68$, $p<0.001$; Appendix Table~\ref{tab:persona_effects}).

\subsection*{Larger Teams Do Not Yield Better Ideas}
Increasing team size from three to six members did not improve creativity for either human or LLM teams. Specifically, for GPT-4.1, larger teams exhibited slightly lower (non-significant) overall creativity ($M_{3}=0.280$ vs. $M_{6}=0.271$, $p=0.219$), driven by increased novelty but decreased usefulness. Human teams also showed lower overall creativity in larger groups ($M_{3}=0.155$ vs. $M_{6}=0.142$, $p=0.188$). These findings indicated that the coordination overhead of larger teams may offset potential creative gains, mirroring classic ``process loss'' effects in organizational psychology \cite{diehl1987productivity} and aligning with recent evidence of coordination failures in AI agent teams \cite{khatua2026cooperbenchcodingagentsteammates} (see Appendix Table~\ref{tab:team_size_llm_human}).

\section*{Semantic Trajectory Analysis Reveals Distinctive Mechanisms of Human and LLM Team Creativity}
To investigate the generative processes underlying both human and LLM team creativity, we developed a semantic trajectory analysis framework. Building on a line of work that analyzes semantic relationships through embedding geometry for cultural meaning~\cite{kozlowski2019geometry}, individual semantic production and semantic navigation~\cite{toro2025semantic, nour2023trajectories}, semantic search and foraging \cite{hills2012optimal,hills2015exploration} and dialogue coherence~\cite{vakulenko2018measuring}, we represented multi-agent conversations as paths through neural embedding space~\cite{mikolov2013word2vec,peters-etal-2018-deep}. This methodology enabled quantitative characterization of exploration breadth, coherence, and convergence dynamics, which are key features underlying the mechanisms of team creativity. Because trajectory analysis requires sequential turn-by-turn dialogue, we analyzed LLM conversations from the \textit{open}, \textit{iterative}, and \textit{instructed} discussion structures (N=3,574) as well as human discussions (N=308). The \textit{Progressive} structure, which uses independent generation followed by collaborative refinement rather than sequential turns, was excluded from trajectory analyses but included in all creativity outcome comparisons. We computed nine trajectory features capturing exploration breadth (path length, semantic spread, max distance), coherence (local and global), and dynamics (convergence ratio, curvature, topic switching rate, revisit score) for each conversation (Table~\ref{tab:trajectory_definitions}; see Methods).

\begin{figure}[ht]
\centering
\begin{subfigure}{0.45\textwidth}
    \includegraphics[width=\textwidth]{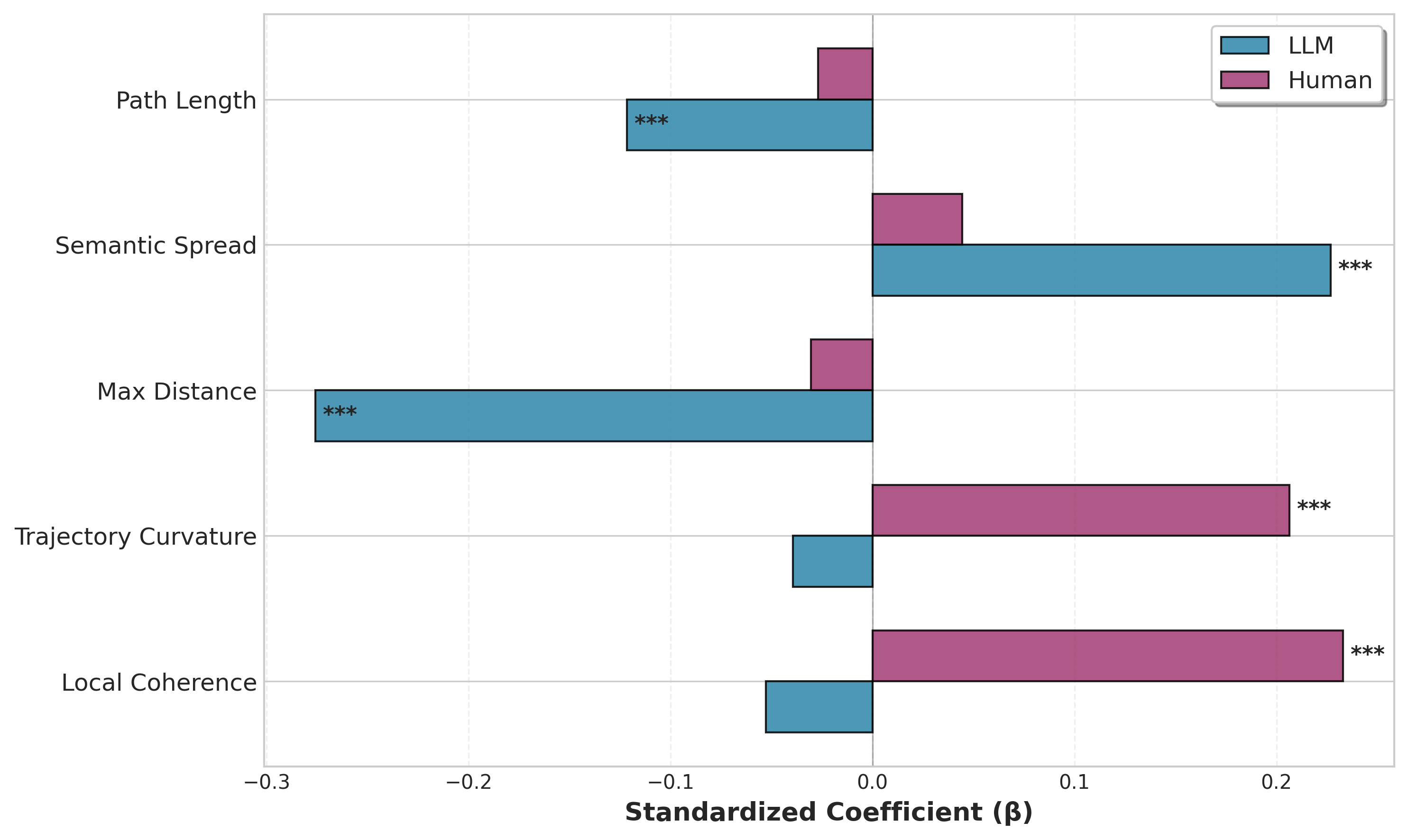}
    \caption{}
    \label{fig:human_llm_different_strategy}
\end{subfigure}
\hfill
\begin{subfigure}{0.45\textwidth}
    \includegraphics[width=\textwidth]{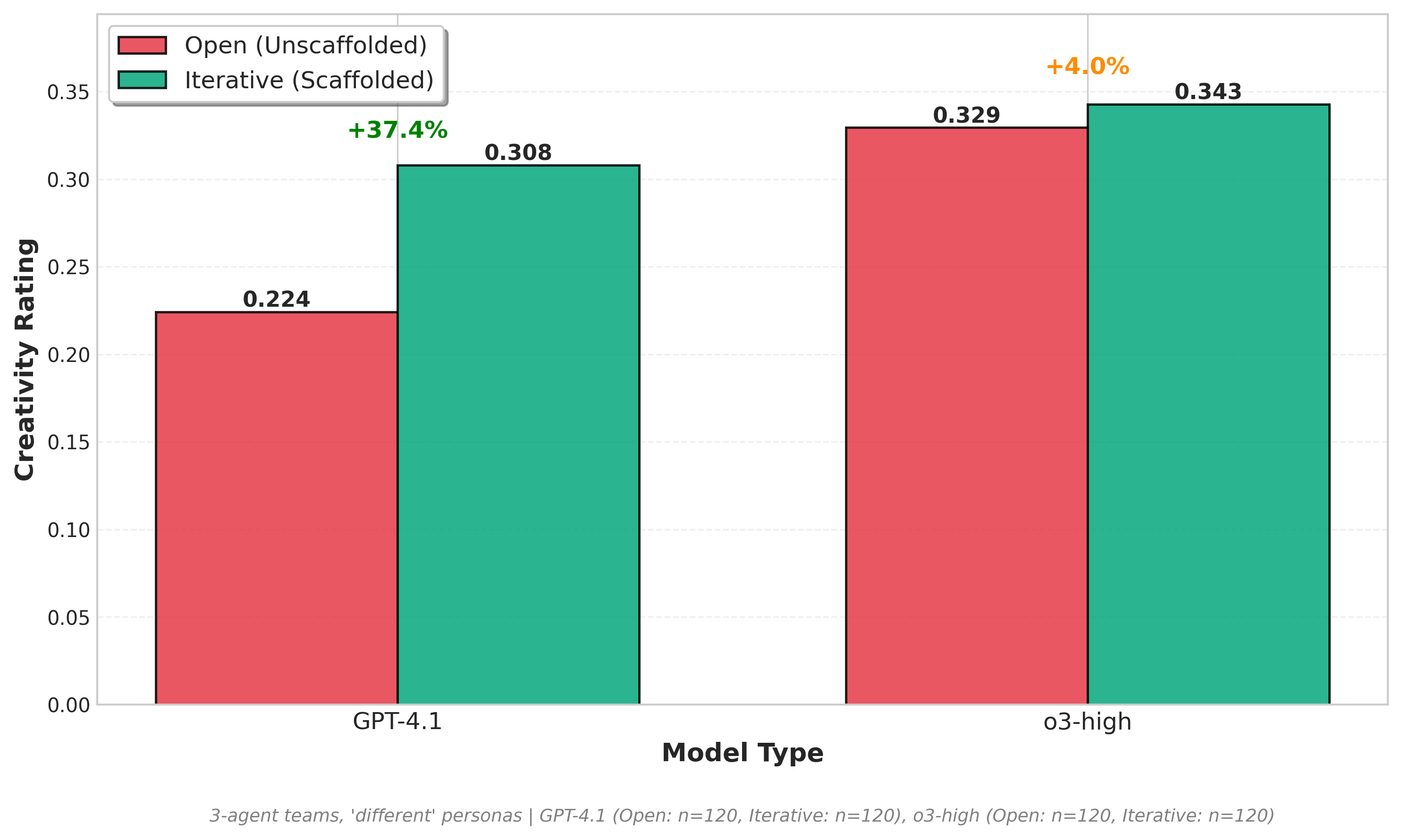}
    \caption{}
    \label{fig:scaffolding_paradox}
\end{subfigure}
\caption{\textbf{Trajectory analysis reveals mechanistic differences between LLM and human teams.} (\textbf{a}) Standardized regression coefficients predicting creativity from trajectory features for LLM teams (blue) and human teams (purple). All features and outcomes are z-scored within sample. Both LLM and human teams produce more creative ideas when conversations range widely (negative global coherence). LLM teams additionally benefit from efficient exploration (positive semantic spread, negative path length and max distance). Human teams additionally benefit from smooth conversational flow (positive local coherence, positive trajectory curvature). For clarity here we only show the five features with the largest absolute LLM-human coefficient differences; all nine features are reported in Table~\ref{tab:llm_human_coefficients}. (\textbf{b}) Scaffolding benefit (open vs. iterative discussion) by model type. GPT-4.1 shows substantial creativity gains from iterative scaffolding (+37.4\%), while o3-high shows minimal benefit (+4.0\%), consistent with reduced dependence on external scaffolding in reasoning models.}
\label{fig:trajectory_mechanisms}
\end{figure}

\subsection*{LLM and Human Teams Differ in Their Semantic Exploration Strategies}

Creative ideas emerged when conversations ranged widely rather than staying centered on a single theme—a pattern observed in both human and LLM teams, as evidenced by negative global coherence coefficients in both groups ($\beta=-0.463$, $p<0.001$ for human teams; $\beta=-0.400$, $p<0.001$ for LLM teams; Figure~\ref{fig:human_llm_different_strategy}; Table~\ref{tab:llm_human_coefficients}). However, the additional trajectory features that predict creativity differ between human and LLM teams.

Human team creativity was associated with maintaining smooth conversational flow: gradual turn-to-turn transitions (local coherence: $\beta=+0.233$, $p<0.001$) and frequent directional pivots (trajectory curvature: $\beta=+0.207$, $p<0.001$). This pattern suggested that human teams achieve wide-ranging exploration through connected pivots—zigzagging across semantic space while preserving local continuity.

In task-adjusted trajectory models, LLM team creativity, by contrast, was associated with efficient exploration: high semantic dispersion (semantic spread: $\beta=+0.227$, $p<0.001$) combined with shorter paths (path length: $\beta=-0.122$, $p<0.001$) and constrained maximum distances (max distance: $\beta=-0.276$, $p<0.001$). Local coherence showed no relationship with LLM team creativity ($\beta=-0.053$, n.s.), suggesting LLM teams can make direct semantic jumps without the discourse constraints that shape human team exploration. These coefficient patterns were robust to alternative embedding models (see Section~\ref{sec:robustness}).

To illustrate these mechanistic differences, we visualized semantic trajectories for two conversations from the same condition (instructed discussion, GPT-4.1, plastic waste task) with contrasting creativity ratings (Figure~\ref{fig:trajectory_examples}). The high-creativity conversation (rating=0.551) exhibited broad semantic exploration (global coherence=-0.91), with dramatic conceptual leaps transforming the idea from economic incentives to blockchain to biotechnology to living organisms. In contrast, the low-creativity conversation (rating=0.090) showed narrow semantic refinement (global coherence=+2.20), iteratively refining a single core concept without venturing into new conceptual territory (Table~\ref{tab:conversation_comparison}).

\subsection*{Design Variables Substantially Shape LLM Conversational Dynamics}

\begin{table}[h]
\centering
\caption{\textbf{Hierarchical modeling shows substantial explanatory power for LLM creativity.} 
Hierarchical OLS models with robust standard errors (HC3), except Mixed Effects which uses REML estimation with task random intercepts. 
Trajectory features, task context, model type, and discussion structure together explain 32.6\% of variance in LLM creativity. 
Human data shown for reference; note that OLS models for LLM teams include predictors (model type, discussion structure variants) without direct human equivalents.
LLM N=3,574; Human N=308.}

\label{tab:trajectory_variance}
\begin{tabular}{lcc}
\hline
\textbf{Model Specification} & \textbf{LLM} & \textbf{Human} \\
\hline
Trajectory features only & 8.6\% & 11.6\% \\
Trajectory + Task & 17.6\% & 15.3\% \\
Trajectory + Task + Model & 22.3\% & — \\
Trajectory + Task + Model + Discussion & 24.4\% & — \\
Mixed Effects (conditional R²) & \textbf{32.6\%} & 17.0\% \\
\hline
\end{tabular}
\end{table}

Hierarchical modeling showed that trajectory features and design variables together explain substantial variance in LLM creativity. A mixed-effects model including trajectory features, task context, model type, and discussion structure accounts for 32.6\% of variance (conditional $R^2$; Table~\ref{tab:trajectory_variance}). This demonstrated that LLM teams' creative performance depends systematically on observable features of conversation structure and experimental design choices. Reasoning models showed substantially higher predictability than standard models: trajectory features explain 46.4\% of variance for o3-high versus 27.9\% for GPT-4.1 (Table~\ref{tab:model_variance}), indicating that reasoning models produce more structured, consistent conversation dynamics.

Given that trajectory features significantly predicted creative outcomes, a key practical question is whether these patterns could be experimentally controlled. Hierarchical regression models revealed that discussion structure and model type together explained 26.8\% of trajectory variance on average across all nine features (Table~\ref{tab:trajectory_manipulability}). The additive model (discussion $+$ model, no interaction) explained 22.8\%, with interaction effects contributing an additional 4.0\%, indicating that optimal discussion methods vary by model type (see Appendix Table~\ref{tab:trajectory_manipulability_detailed} for feature-by-feature breakdown).

\begin{table}[h]
\centering
\caption{\textbf{Design levers explain substantial variance in trajectory features.} 
Hierarchical OLS models with robust standard errors (HC3) predicting trajectory features from experimental design choices. 
Mean R² across nine trajectory features (local coherence, global coherence, path length, convergence ratio, max distance, trajectory curvature, topic switching rate, revisit score, semantic spread). 
N=3,574 LLM conversations.}
\label{tab:trajectory_manipulability}
\begin{tabular}{lc}
\hline
\textbf{Model Specification} & \textbf{Mean R²} \\
\hline
Task only (baseline) & 1.8\% \\
Task + Discussion Structure & 13.3\% \\
Task + Model Type & 11.8\% \\
Task + Discussion + Model (additive) & 22.8\% \\
Task + Discussion × Model (interaction) & 26.8\% \\
\hline
\end{tabular}
\end{table}

Discussion structure primarily determined exploration breadth (e.g., max distance: 40.6\% from discussion vs. 2.7\% from model), while model type primarily affected coherence patterns (e.g., local coherence: 31.4\% from model vs. 11.4\% from discussion; see Table~\ref{tab:trajectory_manipulability_detailed}). Several of the most design-responsive trajectory features were also associated with LLM team creativity, including global coherence, max distance, and semantic spread (Table~\ref{tab:llm_human_coefficients}). Thus, the trajectory framework links manipulable design choices to conversational properties that are associated with creative outcomes, providing a basis for engineering more creative multi-agent systems.

\section*{Discussion}

\subsection*{LLMs as Creative Agents: A Large-Scale Validation}

Our study provides the first large-scale, systematically controlled comparison of LLM and human team creativity across model type, team size, persona assignment, discussion structure, and other conversation features. Across six applied creative problem-solving tasks, LLM teams substantially outperformed human teams ($M=0.297$ vs. $M=0.151$, $d=1.50$), suggesting that the observed advantage is not confined to a single task or domain. This advantage also extended to the upper tail of the distribution: the top LLM idea exceeded the top human idea by 58\%, and the top 5\% of LLM ideas ($M=0.53$) substantially exceeded the top 5\% of human ideas ($M=0.37$). This upper-tail result is notable because prior work on individual LLM creativity has found a more mixed pattern: LLMs often match or exceed average human performance on divergent-thinking tasks, but the most creative humans can still outperform LLMs~\cite{wang2025largescale,bellemare-pepin_divergent_2024,stevenson_putting_2022,nath_characterising_2024}. Our findings therefore extend the literature beyond constrained divergent-thinking tasks to applied creative problem-solving, where novelty must be integrated with practical usefulness and where the models evaluated here substantially exceeded human team performance.

Several factors likely contribute to this divergence from prior work. First, the tasks differ fundamentally. The Divergent Association Task (DAT) and Alternative Uses Task (AUT) are constrained divergent-thinking exercises: generating distant words or unusual object uses. DAT is explicitly scored by semantic distance, while AUT is traditionally evaluated using human or expert ratings of originality, fluency, flexibility, and related dimensions, with semantic-distance methods increasingly used as automated proxies~\cite{olson2021naming,beaty2021automating}. Our tasks instead require integrated creative problem-solving: generating ideas that are simultaneously novel \textit{and} useful across complex real-world domains, evaluated holistically by blinded human judges using a Consensual Assessment Technique-style procedure~\cite{amabile1982social}. LLMs' extensive training on diverse knowledge domains may confer a particular advantage in associating concepts across fields to produce applied solutions, a strength less relevant when the task is simply to name unrelated words. Second, automated semantic-distance scoring captures only one dimension of creativity, whereas human judges can assess both novelty and usefulness, dimensions where LLMs' capacity for cross-domain synthesis may be especially apparent. Third, we tested more recent models, including GPT-4.1 and o3, that may reflect capability advances over those in many prior studies.

\subsection*{Distinct Conversational Patterns Underlie LLM and Human Team Creativity}

Trajectory analysis reveals distinct conversational predictors of creativity. Both LLM and human teams benefit from thematic deviation (negative global coherence; LLM: $\beta=-0.400$, $p<0.001$; human: $\beta=-0.463$, $p<0.001$), a pattern consistent with what \citet{dual_process_creativity} term ``cognitive flexibility.'' However, the remaining predictive features diverge significantly between the two groups.

For LLM teams, highly creative outcomes are strongly linked to efficient exploration, characterized by high semantic spread ($\beta=+0.227$, $p<0.001$) combined with shorter paths ($\beta=-0.122$, $p<0.001$). This indicates that targeted divergence matters more than extensive wandering. Conversely, human creativity is most strongly associated with discourse coherence. Features such as trajectory curvature ($\beta=+0.207$, $p<0.001$) and local coherence ($\beta=+0.233$, $p<0.001$) suggest a pattern of smooth transitions and frequent directional pivots that maintain conversational flow.

Ultimately, LLM creative outcomes appear more strongly associated with observable semantic geometry, including exploration and convergence patterns measurable directly from text. In contrast, human creativity likely depends on unobserved psychosocial factors beyond semantic trajectories, such as tacit knowledge, social dynamics, embodied intuitions, and serendipitous associations~\cite{amabile1996creativity,edmondson1999psychological,harvey2013creative}. As trajectory features possess substantial predictive power for LLMs, these metrics can inform the optimization of conversational design. This establishes a foundation for actively engineering, rather than merely evaluating, multi-agent creativity.

\subsection*{Design Levers Enable Systematic Control of Conversational Patterns}

Our results identify two experimentally manipulable levers that together shape conversational patterns: model choice and discussion structure. These operate through distinct mechanisms, creating a two-dimensional design space.

\textbf{Model choice} determines the baseline level of internal reasoning. Trajectory features explained 46.4\% of creativity variance for reasoning models (e.g., o3-high)—substantially more than those of standard models (GPT-4.1: 27.9\%). This reflects o3's capacity for extended internal deliberation: reasoning models perform structured exploration without external prompting. The practical implication is that reasoning models require less external scaffolding but show ceiling effects where additional structure provides diminishing returns.

\textbf{Discussion structure} modulates the exploration-exploitation balance. Relative to the no-discussion multi-agent baseline, iterative refinement increased GPT-4.1 creativity by 45.3\% ($M=0.308$ vs. $M=0.212$), but produced smaller gains for o3-high (+6.5\%; $M=0.343$ vs. $M=0.322$), suggesting that internal and external reasoning capabilities may substitute for each other. This aligns with recent evidence that reasoning models perform internal multi-agent-like deliberation \cite{kim2026reasoningmodelsgeneratesocieties}, reducing the benefit of external scaffolding. Discussion structure primarily controlled exploration breadth (max distance: 40.6\% variance explained by discussion vs. 2.7\% by model), while model type primarily controlled coherence patterns (local coherence: 31.4\% by model vs. 11.4\% by discussion).

Together, these levers explained 26.8\% of trajectory variance, demonstrating that conversation dynamics are experimentally controllable design parameters rather than fixed model properties. Several highly controllable features were also linked to creative outcomes: global coherence and max distance were associated with LLM team creativity, while local coherence was more strongly associated with human team creativity. This suggests that design choices can target conversational patterns relevant to creativity.

The core implication is that maximizing team creativity requires pairing the model with a complementary discussion structure: standard models appear to benefit most from sequential, iterative scaffolding to push them into broad semantic exploration, while reasoning models maintain high creativity across both structured and
unstructured protocols (like the \textit{open} condition), consistent with the possibility that their internal reasoning processes reduce the need for external conversational scaffolds.

\subsection*{Trajectory Analysis as a Generalizable Framework}
The semantic trajectory framework provides a quantitative lens for understanding how creative outcomes emerge from conversation dynamics. By representing discussions as paths through embedding space, we operationalized abstract concepts like exploration, coherence, and convergence as measurable geometric features drawing on prior work in semantic memory, discourse coherence, and creative search (see Table~\ref{tab:trajectory_definitions} for definitions and theoretical foundations).

Our findings establish the practical utility of this framework. First, trajectory features predict 32.6\% of creativity variance in LLMs, confirming that discussion structure carries meaningful information about creative outcomes. Second, design choices explain 26.8\% of trajectory variance, confirming that these predictive features are sensitive to experimental intervention. The key insight is not that any individual predictor is controllable in isolation, but that the features most strongly associated with creativity, e.g., global coherence and max distance, are also among the most experimentally responsive, creating a tractable pathway from design decisions to creative outcomes.

Beyond validating existing methods, our framework offers a hypothesis-generating approach for designing new interaction mechanisms. While state-of-the-art LLMs remain black boxes that prohibit direct manipulation of model weights or training procedures, our results demonstrate that multi-agent creativity can be shaped through accessible structural levers.

When deploying new foundation models, it is challenging to predict a priori which discussion structure will maximize creativity. However, the trajectory framework provides a generalizable diagnostic heuristic: practitioners can compute trajectory features from a small set of pilot conversations to diagnose a new model's default semantic behavior, and then select the discussion structure that corrects its geometric deficits. If a baseline model produces narrow, highly coherent paths (low max distance, high global coherence), developers should impose forcing protocols (like iterative refinement) to artificially broaden exploration. Conversely, if a model naturally generates highly divergent leaps, designers should implement structures that help agents synthesize those disparate concepts cohesively.

The complementary findings established here support the practical utility of this approach: trajectory features predict 32.6\% of creativity variance in LLMs, and experimental design choices explain 26.8\% of trajectory variance. By understanding a model's default semantic trajectory, practitioners can rely on quantifiable geometric targets rather than trial-and-error to optimize the exploration-exploitation balance. More broadly, this work offers a form of system-level interpretability into multi-agent AI systems, shedding light on how agent interactions produce creative outputs even as individual models remain inscrutable. As multi-agent systems are increasingly applied to scientific discovery~\cite{lu2026towards,novikov2025alphaevolvecodingagentscientific}, the same trajectory diagnostics could also help assess whether AI research agents explore solution spaces effectively or become trapped in narrow search patterns.

\subsection*{Implications for Creativity Theory}

Our findings contribute to creativity theory and practice in multiple ways. First, we demonstrate that the divergent-convergent cycle \cite{guilford1956structure} can be computationally implemented through multi-agent discussions, with measurable geometric signatures distinguishing exploration from exploitation phases. Second, we provide the first quantitative comparison of human and LLM team-level creative processes using a common metric (semantic trajectories), revealing that trajectory features capture a substantial proportion of the creative process for both LLM and human teams. Third, we extend team creativity research \cite{paulus2000groups,nijstad2006group} to artificial agent collectives, showing that coordination mechanisms and structural interventions transfer from human to machine teams.

\subsection*{Limitations}
Several considerations contextualize the scope of our findings. First, creativity is conceptually broad, encompassing everyday problem-solving, artistic expression, and scientific discovery. We operationalize creativity as the product of novelty and usefulness \cite{runco2012standard}, assessed through the Consensual Assessment Technique \cite{amabile1982social}—the definition that has enabled cumulative progress in creativity research and captures the dimension most relevant to practical applications. This operationalization is well-suited to the problem-solving domain we examine where both originality and feasibility matter. Domains with different evaluative criteria—such as pure aesthetic judgment or paradigm-shifting theoretical insight—may require adapted metrics, though the fundamental relationship between conversation structure and creative outcomes should generalize.

Second, our trajectory framework analyzes text-based conversations through semantic embeddings. This focus enables rigorous quantification and experimental control while addressing problems of immediate practical importance. The underlying principle of representing creative exploration as movement through representational space is domain-general, and the geometric patterns we identify (exploration breadth, coherence, convergence) reflect fundamental cognitive processes. Extension to multimodal creativity (e.g., visual design, musical composition) will require domain-appropriate representational spaces but can build on the mechanistic insights established here.

Finally, our study compares LLM teams to human teams but does not examine human-AI collaboration. Our contribution is foundational: we establish that multi-agent LLM systems achieve high creativity and identify the mechanisms (trajectory patterns) and design levers (model choice, discussion structure) associated with creative outcomes. Human-AI collaboration introduces distinct design considerations including real-time coordination protocols, integration of complementary knowledge, and dynamic task allocation between human and AI team members. Our mechanistic findings that specific trajectory features predict creativity and respond to experimental manipulation provide the empirical foundation for addressing these questions in human-AI hybrid systems.

\subsection*{Conclusion}

In creativity, a domain often considered distinctively human, we observe that multi-agent LLM teams substantially outperform human teams ($d=1.50$), primarily driven by novelty advantages while maintaining comparable usefulness. Based on a novel semantic trajectory analysis framework, we reveal distinct exploration patterns between human and LLM teams: both groups produce more creative ideas when conversations range widely, but LLM teams additionally benefit from efficient exploration (high semantic spread, shorter paths), while human teams additionally benefit from maintaining smooth conversational flow (high local coherence, high curvature). These exploration strategies are shaped by design choices: model selection and discussion structure function as complementary levers that together explain a substantial proportion of trajectory variance. This framework offers a systematic approach for understanding how design decisions influence conversational dynamics and creative outcomes. Rather than relying on trial-and-error experimentation, researchers can use these insights to target specific trajectory features, such as exploration breadth or coherence, to increase creativity. By identifying quantifiable targets and manipulable parameters, this work provides a foundation for more principled design and optimization of multi-agent creative systems.

\section*{Methods}

\subsection*{Human Data Collection}

\textbf{Participants.} We recruited 462 participants (40 teams of six and 74 teams of three) through university and college mailing lists and the SONA Research Management system maintained by Cambridge Judge Business School. Eligibility criteria required participants to be at least 18 years old and fluent in English. Participants received £15 for approximately one hour of participation. Of the 462 participants who took part, 460 provided demographic information. 270 (58.7\%) were female, and 188 (40.9\%) male, and two (0.4\%) identified as other. In terms of ethnicity, 96 (20.9\%) were White (British, Irish, other), 72 (15.7\%) Black/Black British (Caribbean, African, other), 114 (24.8\%) Asian/Asian British (Indian, Pakistani, Bangladeshi, other), 124 (27.0\%) Chinese/Chinese British, 22 (4.8\%) of mixed race, 24 (5.2\%) of other ethnic group, and eight (1.7\%) chose not to reveal ethnicity. Mean age was 27.8 (SD = 6.0, range 18–60). 168 (36.5\%) had or were studying for an undergraduate degree, 255 (55.4\%) a graduate degree, and 28 (6.1\%) a doctorate degree. Eight (1.7\%) had a high school diploma or A-levels and one (0.2\%) chose not to reveal their highest educational level. 249 (54.1\%) were currently students, 82 (17.8\%) unemployed and job-seeking, 52 (11.3\%) full-time employed, 34 (7.4\%) part-time employed, 16 (3.5\%) self-employed, 14 (3.0\%) due to start a new job, one (0.2\%) not in paid work (looking after family or home), one (0.2\%) not in paid work (sick or disabled), five (1.1\%) in other employment status, and six (1.3\%) chose not to reveal employment status. 

There are a total of 342 human ideas, among which 341 are valid (one was removed during rating). 308 (from 103 unique teams) of them have complete recordings of team discussions.

\textbf{Procedure.} When signing up for the experiment, participants formed teams of three or six members. Each team completed one of two task sets comprising three problems each in an random order. Each team was allowed up to 12 minutes per task and required to submit a single solution. Following each session, participants individually completed surveys capturing demographic variables (age, gender, ethnicity, education, employment, work experience, professional area, discipline, etc.) and Big Five personality profiles.

\textbf{Transcription.} Discussions were transcribed using Microsoft Azure's speech-to-text service with speaker diarization. Validation on 12 discussions (1,501 utterances) showed 93.3\% semantic accuracy and 89.7\% speaker diarization accuracy.

\subsection*{LLM Data Collection}

\textbf{Multi-Agent System.} We implemented a multi-agent system using four types of LLM settings: all agents based on GPT-4.1~\cite{openai2025gpt41}, o3-high, or o3-low (o3 with reasoning effort set to high/low)~\cite{openai2025o3o4mini}, and mixed model ensembles combining DeepSeek-R1~\cite{guo2025deepseekr1}, Gemini-2.5-Pro~\cite{comanici2025gemini}, and o3-default. Models were accessed via Azure OpenAI API (GPT-4.1, o3, DeepSeek-R1) and Google AI Studio (Gemini-2.5-Pro). For GPT-4.1 and mixed teams, temperature was set to 1.0; o3 models do not support temperature parameters. Teams comprised three or six agents. GPT-4.1 was tested across all experimental conditions, while o3-high, o3-low, and mixed models were tested with three-agent teams using ``different'' personas. This design choice allowed us to comprehensively map GPT-4.1's behavior while efficiently testing whether reasoning models exhibit qualitatively different patterns.

\textbf{Persona Manipulation.} Agents were assigned one of three persona conditions according to prior work in persona prompting~\cite{hu-collier-2024-quantifying,duan2025powerpersonalityhumansimulation}: (1) \textit{none}: generic agent identity (``You are Agent 1''), (2) \textit{same}: all agents received identical creative persona descriptions drawn from real human participants, (3) \textit{different}: each agent received distinct persona descriptions from different human participants, emphasizing varied creative thinking styles.

\textbf{Experimental Design.} We systematically varied design features including model type, team size (three or six agents), persona assignment (none/same/different), discussion structure (open/instructed/iterative/progressive), and conversation structure (with/without initial generation phase), yielding 71 experimental conditions. The study was pre-registered prior to data collection (registered 2 May 2025; \url{https://aspredicted.org/4tht-m898.pdf}). The pre-registration specified the dependent variable (creativity ratings via Consensual Assessment Technique), independent variables (conversation mechanism, number of agents, persona configuration, discussion length and order, model choice), and planned analyses (t-tests, ANOVA, linear mixed models, and exploratory linguistic feature analysis). The trajectory analysis framework extends the pre-registered exploratory text-analysis plan. See Section~\ref{sec:implementation_detail} for complete implementation details.

\textbf{Discussion Structures.} We tested five protocols operationalizing distinct collaborative mechanisms~\cite{osborn1953applied,paulus2000groups,nijstad2006group}: (1) \textit{Open Discussion}: unstructured brainstorming with a summarization step at the end, (2) \textit{Instructed Discussion}: structured discussion cycles with options to agree/modify/replace ideas from a shared pool generated by the model in the initial idea generation stage, (3) \textit{Iterative Refinement}: proposal-and-selection cycles proposing a single new idea and comparing it to existing ones, (4) \textit{Progressive Improvement}: explicit separation of divergent generation (prioritizing novelty) and convergent refinement (enhancing usefulness), and (5) \textit{None}: no collaborative discussion following initial idea generation and rating. Some conditions included initial independent idea generation and rating before discussion; others proceeded directly to collaborative discussion. 
See Section~\ref{sec:implementation_detail} for detailed implementation of each protocol.

\textbf{Data Collection.} A total of 4,740 ideas were generated. Excluding ideas identified as either identical to other ideas or non-relevant to the problem-solving tasks (receiving at least three zero ratings among the five judges) yielded a total of 4,541 multi-agent ideas and 179 single-agent ideas across six creative problem-solving tasks.

\subsection*{Creative Problem-solving Tasks}

Six creative problem-solving tasks were designed to span diverse domains and systematically address potential memorization concerns (Table~\ref{tab:task_prompts}). Tasks comprised two societal challenges (plastic waste, educational inequality), two business problems (supply chain, employee attrition), and two hypothetical scenarios deliberately unlikely to appear in training data (sorry pandemic, singing in shower). This balanced design tests whether LLM performance stems from generalization and novel combinatorial generation rather than the retrieval of memorized solutions. Empirically, LLM teams exhibit higher creativity across all six tasks (see Results), providing evidence against memorization as the primary mechanism.

\begin{table}[htbp]
\centering
\caption{\textbf{Experimental Task Prompts.} The six tasks span environmental, logistical, hypothetical, social, organizational, and behavioral domains. All prompts concluded with the identical instruction: \textit{``You are collaborating with other team members to come up with one creative idea to [solve the problem]. The idea will be evaluated on its creativity (i.e., it should be both novel and useful).''}}
\label{tab:task_prompts}
\renewcommand{\arraystretch}{1.4} %
\begin{tabularx}{\textwidth}{l X} %
\toprule
\textbf{Task Domain} & \textbf{Core Premise} \\
\midrule
\textbf{Plastic Waste} & Plastic waste is one of the biggest environmental problems of our lifetime. \\
\textbf{Supply Chain} & Vulnerabilities within a supply chain could lead to uncontrolled costs and inefficient delivery schedules. \\
\textbf{Sorry Pandemic} & Imagine a new pandemic has emerged that is transmitted by saying the word ``sorry''. \\
\textbf{Educational Inequality} & Educational inequality is the unequal distribution of academic resources to disadvantaged and marginalised groups. \\
\textbf{Employee Attrition} & Voluntary attrition happens when an employee decides to leave the company, resulting in the reduction of valued talent in the workforce. \\
\textbf{Singing in Shower} & Imagine a new research study discovers that singing in the shower for 20 minutes or more is good for health. \\
\bottomrule
\end{tabularx}
\end{table}

\subsection*{Creativity Evaluation}

To ensure that creativity ratings reflected idea content rather than stylistic differences between human and LLM outputs, all ideas underwent a standardization process before evaluation. Using GPT-4.1 (temperature=0), each idea was paraphrased into a consistent format: third-person perspective, active voice, accessible language, and approximately 100 words. The paraphrasing prompt instructed the model to preserve core meaning while removing stylistic markers such as branded concept names, excessive technical jargon, and transformation language (e.g., ``evolve X into Y''). We developed the harmonization prompt through iterative pilot testing to reduce stylistic cues while preserving each idea's substantive mechanism, and then applied the finalized prompt uniformly to all human and LLM-generated ideas. This harmonization minimized surface-level differences, encouraging judges to evaluate the substance of ideas rather than stylistic cues that might signal their origin. See Section~\ref{sec:idea_harmonization_prompt} for the complete harmonization prompt.

To mitigate presentation bias and maintain evaluator blinding, ideas were presented to judges in a single randomized order that intermixed human and LLM-generated ideas within each task. All five judges evaluated ideas in the same randomized sequence, controlling for presentation order while preserving rating independence across judges.

All solutions (human and LLM-generated) were independently evaluated by five trained human judges on novelty (originality relative to other solutions) and usefulness (practicality and feasibility) using a 10-point scale, following the Consensual Assessment Technique \cite{amabile1982social}. If an idea was considered irrelevant to the task, it would receive a zero rating. Before full annotation, judges rated pilot sets (80 ideas for each task) and participated in calibration sessions to establish shared anchors. Inter-rater reliability was assessed using intraclass correlation coefficient (two-way mixed-effects for consistency average measure $ICC_{3,5}$): Novelty (Mean ICC = 0.83, minimum 0.73), Usefulness (Mean ICC = 0.75, minimum 0.70). All coefficients and the 95\% confidence intervals are provided in Appendix Table~\ref{tab:ICC}. For each task and rating dimension, raw judge scores were min-max normalized to the range $[0,1]$ as
\[
x'_{id} = \frac{x_{id} - \min_{k \in \mathcal{T}(i)} x_{kd}}{\max_{k \in \mathcal{T}(i)} x_{kd} - \min_{k \in \mathcal{T}(i)}
x_{kd}},
\]
where $x_{id}$ is the raw score for idea $i$ on rating dimension $d$ and $\mathcal{T}(i)$ denotes the set of ideas from the same
task as idea $i$. Creativity was then computed as $\text{Creativity} = \text{Novelty} \times \text{Usefulness}$.

\subsection*{Trajectory Analysis}

\textbf{Turn Embeddings.} Each conversational turn was embedded using Qwen3-Embedding-0.6B~\cite{zhang2025qwen3embeddingadvancingtext}, chosen for its balance of semantic fidelity and computational efficiency.

\textbf{Trajectory Analysis Scope.} Semantic trajectory analysis requires sequential conversational turns to compute features like local coherence, path length, and curvature. The \textit{progressive} discussion structure—where agents independently generate ideas followed by a single collaborative refinement phase—does not produce the multi-turn sequential structure necessary for trajectory computation. Consequently, trajectory analyses were conducted on conversations from the \textit{open}, \textit{iterative}, and \textit{instructed} structures, which all feature turn-by-turn dialogue. The \textit{progressive} conversations were included in overall creativity comparisons but excluded from trajectory feature analyses.

\textbf{Trajectory Features.} Motivated by prior work on semantic memory, discourse coherence, semantic foraging, and divergent-thinking assessment, we computed nine trajectory features for each conversation \cite{hills2012optimal,hills2015exploration,troyer1997clustering,vakulenko2018measuring}. These features are not intended as previously validated one-to-one measures of creativity mechanisms; rather, they are study-specific operationalizations of three theoretically motivated dimensions of semantic exploration: coherence (local and global coherence), exploration breadth (path length, semantic spread, max distance), and convergence dynamics (convergence ratio, curvature, topic switching rate, revisit score). Formal definitions are provided in Appendix Table~\ref{tab:trajectory_definitions}.

Trajectory features were standardized before analysis. We assessed multicollinearity by computing variance inflation factors (VIF) for all features. All VIF values were below 10, indicating that multicollinearity does not pose a concern for regression coefficient estimation. We retained all nine features in subsequent analyses.

\subsection*{LLM Usage}

Large language models were used in two distinct ways in this study. First, LLMs were the experimental systems under investigation: GPT-4.1, o3-high, o3-low, DeepSeek-R1, Gemini-2.5-Pro, and mixed-model teams were used to generate the LLM-team conversations and final ideas analyzed in the paper, as described in the Methods. Second, GPT-4.1 was used as a preprocessing tool to harmonize human and LLM-generated ideas before blinded human evaluation; the finalized harmonization prompt is reported in Section~\ref{sec:idea_harmonization_prompt}.

The authors also used LLM-based tools to assist with code drafting, debugging, data-analysis workflow organization, and language editing. LLM-generated assistance was not used as a substitute for human evaluation of creativity; all creativity ratings were provided by trained human judges.
  
\clearpage
\bibliography{sn-bibliography}%

\backmatter

\bmhead{Supplementary information}

Supplementary information is provided below.

\bmhead{Acknowledgements}

We thank the former CJBS Experimental Lab Manager Erdenejargal Erdenechuluun, the CJBS MBA office, and the CJBS EMBA office for their help with participant recruitment. We thank the five human judges who rated all the ideas and Speech Ocean for facilitating this process. The project is funded by the Cambridge Centre for Data-Driven Discovery Early Career Research Seed Fund, the Cambridge Judge Business School Small Research Grant, and the Accelerating Foundation Models Research Initiative of Microsoft. T.H. acknowledges support by the Gates Cambridge Trust (grant OPP1144 from the Bill \& Melinda Gates Foundation). Y.J. is supported by the Program of China Scholarship Council (grant 202306320294). J.H.-O.'s research is supported by OpenAI's grant to the ``AI Progress through the Lens of Predictable AI Ecosystems'' program, which is based at the Leverhulme Centre for the Future of Intelligence at the University of Cambridge. L.S. gratefully acknowledges financial support from Invesco through their philanthropic donation to Cambridge Judge Business School.

\section*{Declarations}
\subsection*{Funding}
The project is funded by the Cambridge Centre for Data-Driven Discovery Early Career Research Seed Fund, the Cambridge Judge Business School Small Research Grant, and the Accelerating Foundation Models Research Initiative of Microsoft. T.H. acknowledges support by the Gates Cambridge Trust (grant OPP1144 from the Bill \& Melinda Gates Foundation). Y.J. is supported by the Program of China Scholarship Council (grant 202306320294). J.H.-O.'s research is supported by OpenAI's grant to the ``AI Progress through the Lens of Predictable AI Ecosystems'' program, which is based at the Leverhulme Centre for the Future of Intelligence at the University of Cambridge. L.S. gratefully acknowledges financial support from Invesco through their philanthropic donation to Cambridge Judge Business School.

\subsection*{Competing Interests}
The authors declare the following competing interests:
H.L. and X.X. are employees of Microsoft Research and hold stock in Microsoft. The project received support and API credits from  Microsoft Research, which had no role in the ideas and research questions, study design, data collection and analysis, decision to publish or preparation of the manuscript.

\subsection*{Ethics Approval}
The project received ethics approval from the Cambridge Judge Business School Research Ethics Review Group (ref: 24-15). All participants provided informed consent before taking part, and all procedures were carried out in accordance with the approved protocol and relevant institutional guidelines.

\subsection*{Data Availability}
All data needed to reproduce the analyses in this paper is available on \href{https://github.com/pitehu/llm_multiagent_creativity}{GitHub}.

\subsection*{Code Availability}
All code needed to reproduce the analyses in this paper is available on \href{https://github.com/pitehu/llm_multiagent_creativity}{GitHub}.

\subsection*{Author Contribution}
T.H., J.H.-O., D.S. and L.S. conceived the project. X.X., D.S. and L.S. secured funding for the project. T.H., Y.J. and L.S. collected the data. H.L. and L.S. managed the creativity evaluation process. T.H. prepared the result analysis and plotting and drafted the manuscript. All authors contributed to the interpretation of the results and edited and revised the manuscript. D.S. and L.S. supervised the project.

\subsection*{Corresponding authors}
Correspondence to David Stillwell or Luning Sun.

\newpage
\begin{appendices}

\section*{Supporting Information}\label{secSI}
\section{Supporting Text}
\subsection{Experimental Design and Implementation Details}
\label{sec:implementation_detail}

\subsubsection{Complete Experimental Design}
This appendix provides comprehensive implementation details for all experimental manipulations described in the Methods section. Table~\ref{tab:conditions} presents the complete factorial experimental design spanning 71 conditions.

\textbf{Conversation Structure.} A variety of conditions were designed following Osborn's original principles \cite{osborn1953applied}: defer judgment, encourage wild ideas, build on others' contributions, and aim for quantity. Some conditions included an initial generation-and-selection phase where agents generated ideas before discussion. In conditions without initial generation, agents proceeded directly to the collaborative discussion phase, developing ideas from scratch during discussion.%

\subsubsection{Discussion Structure Implementation Details}
  Each experimental condition comprised up to two components: an \textit{idea-generation-and-selection phase} and a
  \textit{discussion phase}. The idea-generation-and-selection phase, when present, produced an initial candidate idea and, in some
  conditions, a replacement pool. The discussion phase specified how agents interacted with the current idea: freely brainstorming,
  modifying or replacing the idea under explicit action rules, iteratively proposing and rating alternatives, or separating divergent
  generation from convergent refinement. Thus, ``discussion structure'' refers to the protocol governing agent interaction after, or
  in the absence of, the initial generation-and-selection phase.

\textbf{No Discussion/None.} This condition served as a multi-agent baseline without subsequent collaborative discussion. Agents
first generated candidate ideas in response to the task prompt. In the interactive generation mode, agents could see previously
generated ideas from other agents and were instructed to produce different ideas. In the nominal generation mode, agents generated
ideas independently without seeing others' contributions, mirroring nominal group technique \cite{diehl1987productivity}. Agents
then rated all generated ideas for creativity on a 1--10 scale, explicitly considering both novelty and usefulness. The highest-
rated idea, based on the mean agent rating, was selected as the final team output. No further discussion or refinement occurred.

\textbf{Open Discussion.} This method implemented unstructured brainstorming, where agents were not provided with explicit turn-taking rules or discussion structure beyond the task description. After reaching the maximum turn limit, one randomly selected agent was prompted to synthesize the entire discussion history into a single coherent final idea (80-100 words), mimicking the consensus-building phase typical of human brainstorming sessions.

\textbf{Instructed Discussion.} This condition imposed an explicit revise-or-accept structure on the conversation. The discussion began either from the top-ranked idea selected during the initial generation-and-selection phase or, in conditions without that phase, from an idea generated at the start of discussion. At each turn, the active agent evaluated the current idea and selected one of three actions:
\begin{itemize}
\item \textbf{Agree}: Accept the current idea without changes (response: ``Agree: No changes needed''). In some conditions, agreement was disabled until a minimum number of discussion rounds (30) had elapsed, forcing extended exploration.
\item \textbf{Modify}: Propose a substantially revised version of the current idea, with an explicit requirement for major improvements rather than minor edits (response format: ``Modify: [full revised idea] - Reason: [justification]'').
\item \textbf{Replace}: Substitute the current idea with an alternative either from a replacement pool or newly generated by the agent (response format: ``Replace: [full replacement idea] - Reason: [justification]'').
\end{itemize}

The replacement pool varied by condition. In no-pool conditions, agents generated a new replacement idea when choosing to replace.
In pool conditions, the five next-highest-ranked ideas from the initial generation-and-selection phase formed a shared replacement
pool: the top-ranked idea became the initial current idea, while ideas ranked 2--6 were available as replacements. Once a pooled
idea was used, it was removed to prevent reuse. If the pool was exhausted, agents generated new replacement ideas. The instructed
discussion ended when all agents agreed on the current idea or when the turn limit was reached.

\begin{itemize}
\item \textit{No pool} (size = 0): If replacement was attempted, agents were instructed to generate an entirely new idea rather than selecting one from the pool.
\item \textit{Replacement pool} (size = 5): In some conditions, the top-rated five ideas (ranked 2-6, excluding the top-ranked idea which became the first current idea) formed a shared replacement pool accessible to all agents. When an agent replaced the current idea with one from this pool, the used idea was removed from the pool, preventing reuse. If all ideas from the replacement pool were used, the agent would be asked to generate a new idea
\end{itemize}

The discussion concluded when all agents agreed on the current idea or the turn limit was reached.

  \textbf{Hand-Raising Mechanism.} A subset of GPT-4.1 instructed-discussion conditions tested an alternative turn-taking rule
  inspired by human meeting dynamics. Instead of using a fixed or random speaking order, agents self-selected based on their
  perceived ability to contribute. At each turn, agents who had not yet agreed rated their desire to speak on a 1--7 scale according
  to how much value their potential contribution would add. The agent with the highest self-rated desire spoke next.

\textbf{Iterative Refinement.} This method implemented iterative refinement with backtracking capability \cite{cropley2006praise}. Unlike instructed discussion, which only allowed replacement from a pool of alternatives, iterative refinement maintained a constrained memory of recently explored ideas.

The process began with the initial idea, which could be newly generated by the starting agent or the top-ranked idea from the initial generation-and-selection phase. At each turn:
\begin{enumerate}
\item One agent generated a new candidate idea intended to be more creative than existing ideas.
\item All agents rated the new candidate, the current idea, and recently explored past ideas on creativity (1-10 scale).
\item The highest-rated idea became the new current idea for the next iteration. This allowed the system to either adopt the new modification, retain the current idea, or backtrack to a recently considered alternative.
\end{enumerate}

This proposal-and-evaluation cycle continued until one of two stopping criteria was met:
\begin{itemize}
\item \textit{Convergence}: The same idea was selected (rated highest) for three or six consecutive rounds depending on the team size, indicating the team had reached a consensus.%
\item \textit{Turn limit}: No convergence condition was reached despite reaching the turn limit.
\end{itemize}

\textbf{Progressive improvement.} This method explicitly separated divergent and convergent thinking phases \cite{guilford1956structure}, operationalizing the theoretical distinction between exploration (generating novel possibilities) and exploitation (refining promising candidates). After the initial generation-and-selection phase, two further phases were implemented:

\textit{Phase 1 - Divergent Generation}: Each agent generated 5 radically novel ideas inspired by, but distinct from, the top-ranked ideas from the initial generation-and-selection round. Agents were explicitly instructed to prioritize novelty over usefulness, to think transformatively about fundamental shifts in approach, and to avoid simply combining existing ideas. 

\textit{Phase 2 - Convergent Refinement}: Agents rated all ideas on novelty (not overall creativity), and the top-ranked novel ideas advanced to a refinement phase. Each agent then refined these ideas to enhance usefulness while carefully preserving the core novelty. 

This two-phase structure tests whether explicitly separating exploration and exploitation yields higher creativity than methods that interleave these processes.

\subsubsection{Conversation Length and Termination}

Conversation length varied systematically across conditions:
\begin{itemize}
\item \textit{30 turns}: Standard length for most conditions, balancing exploration time with computational cost.
\item \textit{60 turns}: Extended length for open discussion conditions, testing whether longer conversations enable more complex semantic trajectories.
\item \textit{Minimum thresholds}: Some instructed discussion conditions enforced a minimum of 30 discussion rounds before agreement was permitted, preventing premature convergence.
\end{itemize}

\textbf{Discussion Order and Turns.} Discussion order was either fixed (predetermined rotating sequence), random (stochastic agent selection at each turn), or based on the hand-raising mechanism.%

\subsection{Idea Harmonization Protocol}
\label{sec:idea_harmonization_prompt}
To ensure stylistic uniformity across human- and LLM-generated ideas before human evaluation, all ideas were paraphrased using GPT-4.1 (temperature=0). The complete system prompt used for harmonization is provided below:
\begin{promptbox}{System Prompt}
\begin{lstlisting}
You are an AI assistant that rephrases ideas. Your main goal is to accurately convey the core meaning of the provided text in a natural, accessible way.

**Text Processing Instructions:**
- If the text contains "evolve [X] into [Y]" or similar transformation language, focus only on the final concept (Y) and ignore the initial concept (X).
- If the text contains multiple ideas with clear section headers (like "---Final Idea---" or "---One Creative Idea---"), focus only on the content under the final/main idea section.
- If the text contains multiple ideas without clear hierarchy, treat the entire content as one comprehensive idea.
- Remove any brainstorming artifacts, transitional phrases, or development language that doesn't contribute to the core concept.

**Writing Style Requirements:**
- Use a **clear, neutral, and straightforward writing style** regardless of the original text's style.
- Simplify complex or technical language while preserving core meaning - choose everyday words over specialized terminology when possible.
- Remove branded terms, metaphorical names in quotes, and elaborate descriptive phrases that don't add essential information.
- Break down dense technical concepts into simpler, more digestible explanations.
- Be concise: If the original idea is lengthy, shorten it. If it's already brief, rewrite while preserving its essence and approximate length.
- Stick to the original meaning: Do not add new information, context, or opinions.
- Direct output: Provide only the rephrased idea, without introductory or concluding phrases.
- Use THIRD-PERSON perspective only. Never use "we", "I", "you", "us", or "our" but maintain natural language.
- Use active voice when possible and replace overly formal or technical language with clear, accessible alternatives.
- Reduce technical density: If the original lists many technical specifications or components, focus on the main mechanism and primary benefits.
- Use conversational sentence structure rather than academic or highly technical phrasing.
\end{lstlisting}
\end{promptbox}

\begin{promptbox}{User Prompt}
\begin{lstlisting}
Paraphrase this idea in under 100 words using simple, clear language that anyone can understand: {idea}
\end{lstlisting}
\end{promptbox}

This preprocessing was applied identically to all ideas prior to human evaluation.

\subsection{Supplementary Analyses: Discussion Structure, Persona, and Team Size Effects}

\subsubsection{Overview}

This appendix provides detailed statistical analyses supporting the main paper findings on discussion structure, persona assignment, and team size effects. We present: (1) full regression results with hierarchical model specifications, (2) novelty-usefulness decomposition revealing underlying mechanisms, (3) descriptive statistics for all model $\times$ discussion combinations, and (4) matched-condition analyses for persona and team size effects.

\subsubsection{Regression Strategy}

To properly account for the unbalanced experimental design (GPT-4.1 tested on all 71 conditions; other models tested only on three-agent teams with ``different'' persona assignments), we restricted the main regression to these common conditions. This ensures balanced comparisons across all models while avoiding confounding effects between model type and experimental factors.

Table~\ref{tab:regression_full} presents four hierarchical OLS models with robust standard errors (HC3). Model 1 includes discussion structure only; Model 2 adds model type; Model 3 adds task fixed effects; Model 4 adds the critical model $\times$ discussion interaction. The interaction model (Model 4) is our primary specification, showing how discussion effects vary by model type. Note that we do not include idea generation mode or discussion order as controls in the primary model because their reference categories (``none'' and ``not applicable'') only occur when those phases are absent, confounding phase presence with phase configuration. Separate analyses examine these factors within relevant subsets (see below).

\subsubsection{Main Finding: Model $\times$ Discussion Interaction}

Table~\ref{tab:discussion_effects} summarizes discussion structure effects by model type from Model 4. For GPT-4.1, all discussion structures significantly improved creativity compared to no discussion (iterative: $\beta=+0.097$, $p<0.001$; progressive: $\beta=+0.084$, $p<0.001$; instructed: $\beta=+0.065$, $p<0.001$; open: $\beta=+0.013$, $p=0.021$). For o3-high and o3-low, most discussion structures showed minimal effects (most $p>0.05$), with only instructed showing marginal significance for o3-high ($\beta=+0.023$, $p=0.028$). The Mixed model (DeepSeek-R1, Gemini-2.5-Pro, o3 ensemble) showed negative main effects for structured discussions, suggesting discussion may interfere with these models' internal reasoning processes.

\subsubsection{Mechanism: Novelty-Usefulness Trade-off}

Table~\ref{tab:novelty_usefulness} presents parallel regressions for novelty and usefulness as separate outcomes, revealing the mechanism behind discussion effects. For the reference model (Mixed), discussion structures that boosted novelty systematically reduced usefulness:
\begin{itemize}
\item Progressive: $\beta_{\text{novelty}}=+0.064$, $p<0.001$; $\beta_{\text{usefulness}}=-0.084$, $p<0.001$
\item Instructed: $\beta_{\text{novelty}}=+0.027$, $p=0.01$; $\beta_{\text{usefulness}}=-0.066$, $p<0.001$
\item Iterative: $\beta_{\text{novelty}}=+0.044$, $p<0.001$; $\beta_{\text{usefulness}}=-0.053$, $p<0.001$
\item Open: $\beta_{\text{novelty}}=-0.068$, $p<0.001$; $\beta_{\text{usefulness}}=+0.021$, $p=0.13$
\end{itemize}

This trade-off explains why GPT-4.1's iterative discussion was most effective: it markedly boosted novelty ($\beta_{\text{interaction}}=+0.208$, $p<0.001$) without proportionally reducing usefulness ($\beta_{\text{interaction}}=-0.036$, $p=0.25$), resulting in the largest net creativity gain.

By contrast, o3 models showed the opposite pattern for iterative discussion: slightly decreased novelty but increased usefulness (o3-high: $\beta_{\text{usefulness}}=+0.069$, $p=0.025$; o3-low: $\beta_{\text{usefulness}}=+0.064$, $p=0.048$). This suggests reasoning models already generate highly novel ideas internally and benefit more from usefulness-focused refinement.

\subsubsection{Descriptive Statistics}

Table~\ref{tab:descriptive_stats_discussion_methods} provides means and standard errors for creativity, novelty, and usefulness across all model $\times$ discussion combinations in common conditions. These descriptive statistics complement the regression analyses and allow readers to verify the patterns reported in the main text.

Key observations:
\begin{itemize}
\item GPT-4.1 baseline (no discussion) shows low creativity ($M=0.212$) driven by low novelty ($M=0.313$), indicating this model requires external structure to generate creative ideas.
\item o3 models maintain high creativity across all discussion structures ($M=0.307$--$0.345$), with baseline performance ($M=0.322$--$0.331$) already exceeding GPT-4.1's best condition.
\item The novelty-usefulness trade-off is visible in the descriptive statistics: iterative discussion consistently shows highest novelty but lowest usefulness within each model.
\end{itemize}

\subsubsection{Distribution of Top Ideas by Source}
To examine whether LLM superiority extends to the highest-performing ideas within each task, we identified the top 10 ideas ranked by creativity, novelty, and usefulness separately for each task (Table~\ref{tab:top10_distribution}). LLM teams accounted for most top-ranked ideas across all dimensions: for creativity, 60 of 60 top-10 ideas (100\%) were LLM-generated. This pattern held for novelty (48/60, 80.0\%) and was weaker but still substantial for usefulness (45/60, 75.0\%). This indicates that LLM superiority is not driven by a few exceptional outliers but reflects a shift in the entire quality distribution.
Table~\ref{tab:top_ideas} presents the actual content of the top 5 ideas per task, illustrating the nature of high-creativity LLM outputs. These ideas demonstrate diverse approaches including technological innovation, behavioral interventions, and systematic redesign.

\subsubsection{Task-Level Novelty-Usefulness Distributions}
\begin{figure}[h]
\centering
\includegraphics[width=0.95\textwidth]{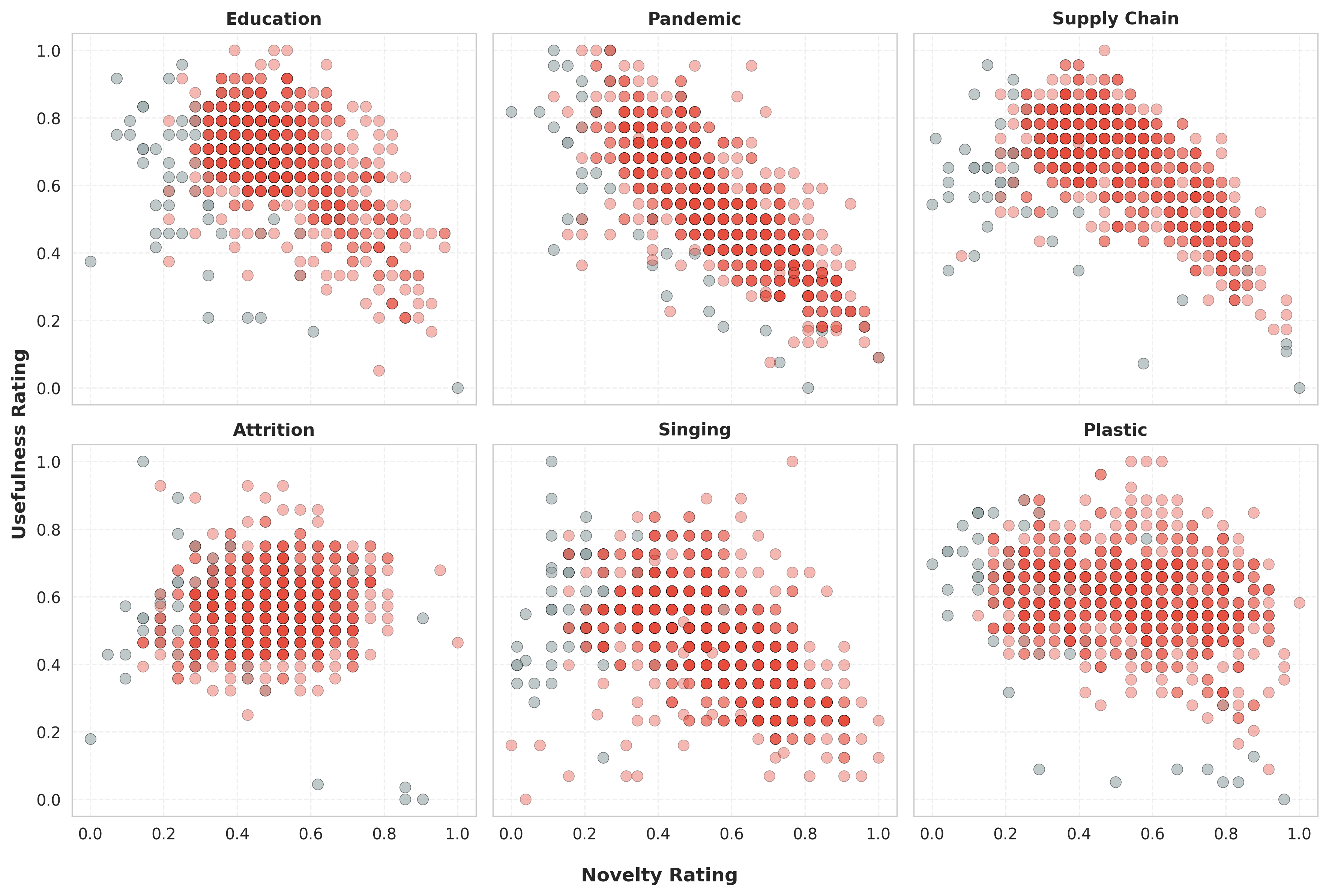}
\caption{\textbf{Novelty-usefulness distributions by task.} 
Each panel shows ideas in novelty-usefulness space for one task (gray = human, red = LLM). 
Human ideas occasionally reach high novelty or high usefulness individually, but rarely achieve both simultaneously. 
LLM ideas more frequently occupy the upper-right quadrant where both dimensions are high, explaining their creativity advantage.}
\label{fig:novelty_usefulness_by_task}
\end{figure}

Figure~\ref{fig:novelty_usefulness_by_task} shows the novelty-usefulness distributions separately for each task. While human ideas (gray) occasionally achieve high novelty or high usefulness individually, they rarely achieve both simultaneously: the upper-right quadrant is sparsely populated by human ideas across all tasks. In contrast, LLM ideas (red) frequently occupy this high-novelty, high-usefulness region. This pattern explains the creativity gap: creativity is computed as the product of novelty and usefulness, so ideas must excel on \textit{both} dimensions to achieve high creativity scores. Human teams face a novelty-usefulness trade-off that LLM teams largely overcome.

\subsubsection{Persona Effects (GPT-4.1 Only)}

Table~\ref{tab:persona_effects} presents condition-matched comparisons for persona assignment effects. Since persona variations only exist for GPT-4.1, this analysis is restricted to GPT-4.1 teams with same, different, or no assigned personas, controlling for all other experimental factors by matching condition codes.

We found no significant omnibus difference in overall creativity (one-way ANOVA: $F=2.09$, $p=0.124$). Different-persona teams performed comparably to no-persona teams ($M=0.267$ vs.\ $M=0.269$; two-sided $t(958)=0.25$, $p=0.803$), whereas same-persona teams showed a small, marginally lower creativity score than no-persona teams ($M=0.258$ vs.\ $M=0.269$; two-sided $t(956)=1.91$, $p=0.057$). However, persona assignment shifted the novelty-usefulness balance:\begin{itemize}
\item No persona: High novelty ($M=0.541$), low usefulness ($M=0.534$)
\item Same persona: Lower novelty ($M=0.461$, $t(956)=7.15$, $p<0.001$), higher usefulness ($M=0.587$,
  $t(956)=-4.98$, $p<0.001$), relative to no persona
\item Different persona: Lower novelty ($M=0.480$, $t(958)=5.31$, $p<0.001$), higher usefulness ($M=0.585$,
  $t(958)=-4.68$, $p<0.001$), relative to no persona
\end{itemize}

This pattern suggests that explicit persona assignments may constrain agents toward more ``realistic'' or role-consistent solutions, reducing exploration of novel ideas while improving practical applicability. The net effect on overall creativity is null because the multiplicative metric appropriately captures the trade-off between these dimensions.

\subsubsection{Team Size Effects (GPT-4.1 and Human)}

Table~\ref{tab:team_size_llm_human} compares 3-agent versus 6-agent teams for GPT-4.1 (matched conditions with different personas assigned to the agents) and humans. Both populations exhibited process loss, though the effect was not statistically significant at conventional levels.

For GPT-4.1:
\begin{itemize}
\item Creativity: $M_{3}=0.280$ vs. $M_{6}=0.271$, $\Delta=-0.009$, $t(716)=1.23$, $p=0.219$, $d=0.09$
\item Novelty: $M_{3}=0.479$ vs. $M_{6}=0.511$, $\Delta=+0.032$, $p<0.001$ (larger teams more novel)
\item Usefulness: $M_{3}=0.606$ vs. $M_{6}=0.558$, $\Delta=-0.048$, $p<0.001$ (larger teams less useful)
\end{itemize}

For humans:
\begin{itemize}
\item Creativity: $M_{3}=0.155$ vs. $M_{6}=0.142$, $\Delta=-0.013$, $t(339)=1.32$, $p=0.188$, $d=0.14$
\item Novelty: $M_{3}=0.320$ vs. $M_{6}=0.287$, $\Delta=-0.033$, $p=0.094$
\item Usefulness: $M_{3}=0.553$ vs. $M_{6}=0.600$, $\Delta=+0.047$, $p=0.048$
\end{itemize}

In particular, the mechanisms differ between LLM and human teams: GPT-4.1 teams show increased novelty but decreased usefulness with larger size, whereas human teams show decreased novelty but increased usefulness. This suggests different coordination challenges: LLM teams may suffer from ``too many ideas'' reducing consensus on practical solutions, while human teams face classic motivation and coordination losses.

\subsubsection{Replacement pool}
We tested whether providing agents with a 
pool of five pre-generated alternative ideas during instructed discussion 
affected creative outcomes. In matched conditions differing only in pool 
availability (conditions 41--45 vs.\ 31--35; Table~\ref{tab:replacement_pool}), 
the replacement pool had no effect on creativity ($d=-0.03$, $p=0.68$), 
novelty ($d=-0.06$, $p=0.40$), or usefulness ($d=0.02$, $p=0.75$). 
This pattern was consistent across all four model types. This null result 
suggests that creative outcomes in instructed discussion are driven by the 
iterative discussion process itself rather than access to a larger set of 
candidate ideas.

\subsubsection{Additional Experimental Factors}

\textbf{Task effects.} Creativity varied substantially by task (see Table~\ref{tab:regression_full}), with ``singing in shower'' showing lowest creativity and ``education inequality'' showing highest, consistent with task difficulty and domain familiarity differences.

\textbf{Discussion order.} Among conditions with discussion, we analyzed whether discussion order (fixed rotation vs. random selection vs. hand-raising) affected creativity (Table~\ref{tab:discussion_order_effects}). Within 3-agent teams with ``different'' personas and discussion phases (N=2,983), the hand-raising mechanism reduced creativity relative to random ordering ($\beta=-0.015$, $p=0.003$), while fixed ordering showed no significant effect ($\beta=-0.003$, $p=0.40$). Within instructed discussions specifically (N=1,785), where all models had all three order conditions, the hand-raising effect remained ($\beta=-0.014$, $p=0.009$). This suggests that the hand-raising mechanism may reduce exploration by favoring confident agents, hence undermining team-level creativity.

\textbf{Idea generation mode.} Among no-discussion conditions (N=240 with three-agent ``different'' persona teams), nominal brainstorming (independent generation) reduced creativity compared to interactive generation ($\beta=-0.033$, $p=0.008$; Table~\ref{tab:generation_effects}), driven primarily by reduced novelty. This indicates that exposure to others' ideas during generation, rather than sheer quantity of independent ideas, drives creative outcomes.

\textbf{Discussion length.} We tested whether discussion length (30 vs. 60 for open discussion and 30 vs. 60 with minimum 30 rounds for instructed discussion) affected creativity, controlling for discussion structure, model, persona, team size, and task (N=3,702; Table~\ref{tab:round_effects}). Results showed no significant effect of discussion length (LR test $p=0.72$, $\Delta R^2 < 0.001$), indicating that discussion protocol quality matters more than duration.

\subsection{Trajectory Analysis}

We analyzed semantic trajectories by computing turn-level embeddings for all conversations using Qwen3-Embedding-0.6B and extracting nine trajectory features capturing exploration dynamics, coherence, and refinement patterns. This appendix provides complete technical details, regression results, and illustrative examples.

\subsubsection{Trajectory Features}

Table~\ref{tab:trajectory_definitions} defines all trajectory features used in our analysis. These features capture three key aspects of semantic exploration: \textit{coherence} (how connected ideas are), \textit{exploration} (how broadly the conversation searches semantic space), and \textit{dynamics} (how the trajectory evolves over time). All features are grounded in prior work on semantic memory~\cite{kenett2014investigating}, discourse coherence~\cite{elvevag2007quantifying}, and creative search~\cite{hills2015exploration}.

\subsubsection{Regression Results}

Tables~\ref{tab:llm_regression_full} and~\ref{tab:human_regression_full} present full regression results for LLM and human team creativity, respectively. We fit hierarchical models to isolate the contribution of trajectory features while controlling for task, model type, and discussion structure. For LLM teams, trajectory features explain 32.6\% of creativity variance (conditional R²), with global coherence and path length as the strongest predictors. For human teams, trajectory features explain 17.0\% of variance, with trajectory curvature and local coherence as key predictors.

Table~\ref{tab:llm_human_coefficients} directly compares coefficients, revealing both shared and distinct mechanisms: Both groups benefit from broad semantic exploration (negative global coherence), but differ in their additional strategies—LLMs benefit from efficient exploration (high semantic spread, low path length), while humans benefit from smooth iterative refinement (high curvature, high local coherence).

\subsubsection{Model-Specific Variance Decomposition}

Trajectory features explained different amounts of variance across LLM model types (Table~\ref{tab:model_variance}). Substantially more variance was explained in reasoning models (o3-high: 46.4\%, o3-low: 36.1\%) than in standard models (GPT-4.1: 27.9\%), suggesting that reasoning models produce more structured and predictable conversation dynamics. The mixed model condition showed the lowest explained variance (19.1\%), consistent with less structured generation patterns.

\subsubsection{Experimental Manipulability of Trajectory Features}
To assess whether trajectory features can be experimentally controlled, we regressed each trajectory feature on discussion structure and model type (controlling for task). Hierarchical regression models revealed that discussion structure and model type together explained 26.8\% of trajectory variance on average across all nine features (Table~\ref{tab:trajectory_manipulability_detailed}). The additive model (discussion + model, no interaction) explained 22.8\%, with interaction effects contributing an additional 4.0\%, indicating that optimal discussion structures vary by model type.
Highly manipulable features included max distance, global coherence, and local coherence, for which the full design model explained 51.7\%, 52.6\%, and 43.4\% of feature variance, respectively. Discussion structure primarily controlled exploration breadth (e.g., max distance: 40.6\% from discussion vs. 2.7\% from model), while model type primarily controlled coherence patterns (e.g., local coherence: 31.4\% from model vs. 11.4\% from discussion). The interaction model outperforms the additive model, with features like semantic spread showing substantial interaction effects (13.6\% additional variance, 55\% of total; Table~\ref{tab:trajectory_interaction_effects}). This supports an effective pathway: design choices explain 26.8\% of trajectory variance, and trajectory patterns predict 32.6\% of creativity variance, demonstrating that conversation structure is both experimentally manipulable and associated with creative outcomes.

\subsubsection{Illustrative Examples}

\begin{figure}[hp]
\centering

\begin{subfigure}[t]{0.48\textwidth}
    \centering
    \includegraphics[width=\textwidth]{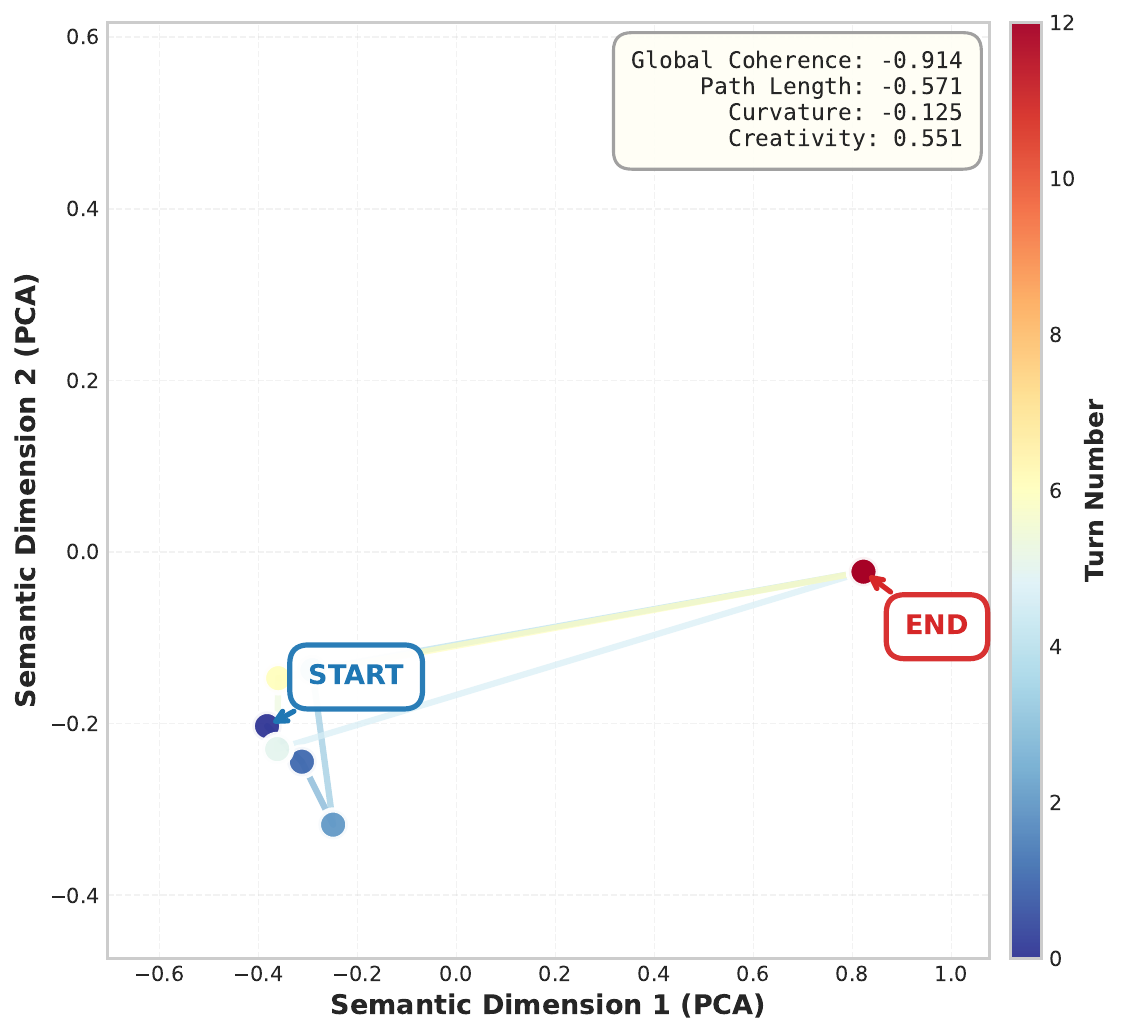}
    \caption{High creativity conversation (rating = 0.551). Trajectory shows broad semantic exploration (global coherence = -0.91), with dramatic conceptual leaps from economic incentives → blockchain → biotechnology → living organisms.}
    \label{fig:traj_high}
\end{subfigure}
\hfill
\begin{subfigure}[t]{0.48\textwidth}
    \centering
    \includegraphics[width=\textwidth]{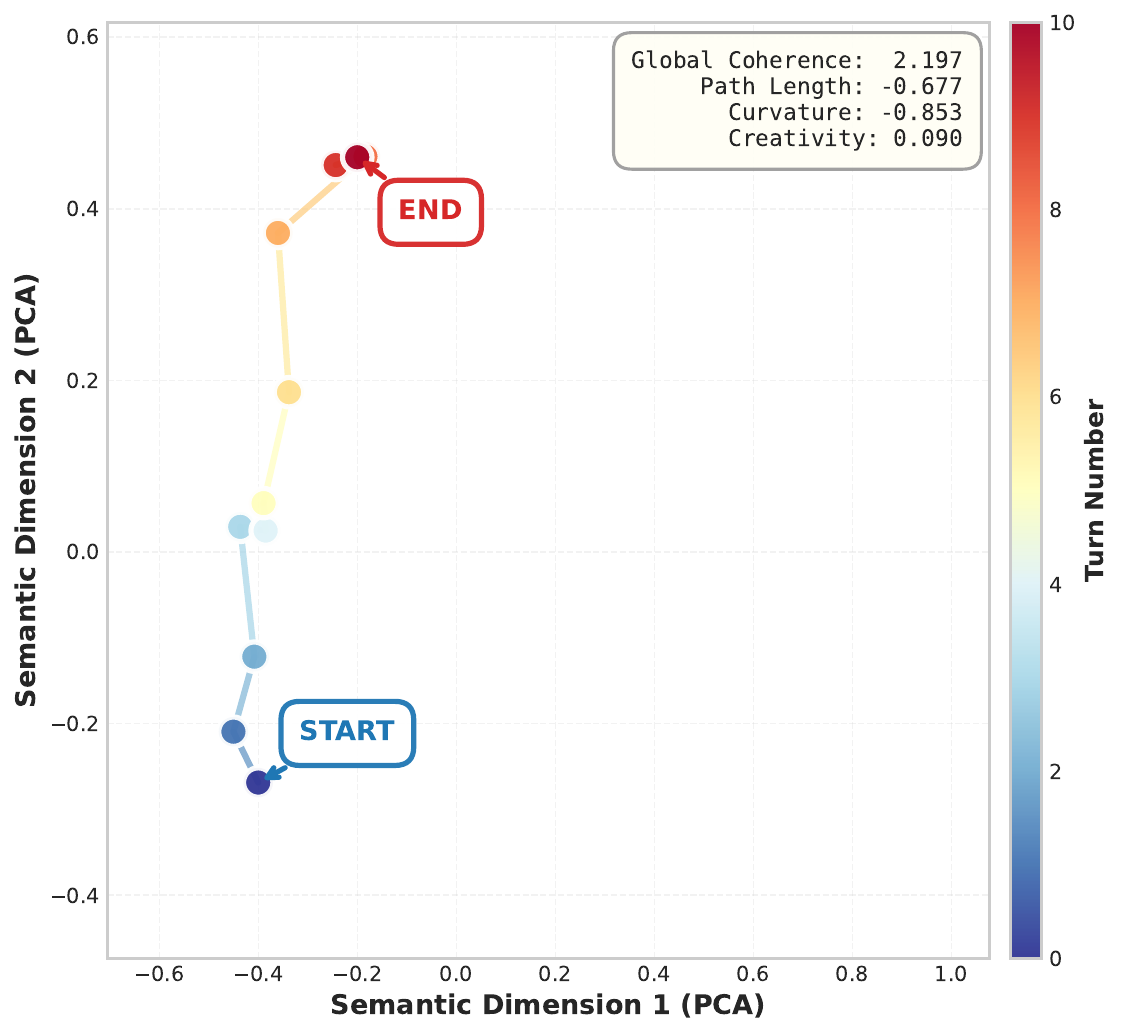}
    \caption{Low creativity conversation (rating = 0.090). Trajectory shows narrow semantic refinement (global coherence = +2.20), iteratively scaling a single core concept (competitive plastic reduction events).}
    \label{fig:traj_low}
\end{subfigure}
\caption{\textbf{Semantic trajectory examples illustrating mechanistic differences in creative exploration.} 
Each panel shows a conversation from the same condition (instructed discussion, GPT-4.1, plastic waste task) projected onto 2D semantic space via PCA. 
Colors indicate temporal progression (blue = early turns, red = late turns). 
The high-creativity conversation (A) explores widely across semantic domains, while the low-creativity conversation (B) remains tightly clustered around a single concept. 
Global coherence (distance from semantic centroid) captures this difference: negative values indicate broad exploration, positive values indicate narrow focus. 
See Table~\ref{tab:conversation_comparison} for full conversation excerpts.}
\label{fig:trajectory_examples}
\end{figure}

Figure~\ref{fig:trajectory_examples} and Table~\ref{tab:conversation_comparison} illustrate these mechanistic differences through two conversations from the same condition (instructed discussion, GPT-4.1, plastic waste task). The high-creativity conversation (rating = 0.551) exhibits dramatic conceptual leaps—from economic incentives to blockchain to biotechnology to living organisms—resulting in broad semantic exploration (global coherence = -0.91). In contrast, the low-creativity conversation (rating = 0.090) iteratively refines a single core concept (competitive plastic reduction events), staying tightly clustered in semantic space (global coherence = +2.20). These examples demonstrate how trajectory features capture the \textit{process} by which creative ideas emerge through multi-turn multi-agent conversations.%

\subsection{Robustness Checks}
\label{sec:robustness}

\subsection*{Robustness to nested idea structure}
\label{sec:robustness_clustered}
The primary LLM-human comparison is reported at the idea level because creativity ratings were assigned to individual ideas. However, ideas are not fully independent: they are nested within tasks and within generating teams or system conditions. We therefore estimated robustness models that account for this structure.

First, we fit an ordinary least squares model predicting normalized product-based creativity from source condition (LLM versus human), including task fixed effects and heteroskedasticity-robust standard errors. The LLM advantage remained large, $\beta=0.146$, 95\% CI [0.135, 0.156], $p<0.001$.

Second, we repeated the task-adjusted model with standard errors clustered by generating cluster, defined as the LLM
condition-by-task cluster for LLM ideas and the human team/session identifier for human ideas. This produced a similar estimate, $\beta=0.146$, 95\% CI [0.134, 0.157], $p<0.001$.

Third, we aggregated to task-level LLM-human differences as a conservative check. The LLM advantage was positive for all six tasks, with a mean task-level difference of $0.148$, $t(5)=8.68$, $p<0.001$.

Finally, a mixed-effects model with random effects for task and generating cluster also supported the same conclusion, $\beta=0.132$, 95\% CI [0.086, 0.179], $p<0.001$. Thus, the main LLM-human difference was robust to analyses that relax the assumption of independent idea-level observations.

\subsubsection{Embedding Model Robustness} To ensure our trajectory metrics were not artifacts of the specific embedding model, we re-computed all semantic trajectory features using a substantially larger model (Qwen3-Embedding-4B, 4 billion parameters) compared to the Qwen3-Embedding-0.6B model used in the main analysis. Table~\ref{tab:robustness_embedding} shows that while explained variance decreased modestly (LLM: 17.6\% $\rightarrow$ 14.7\%; Human: 15.3\% $\rightarrow$ 13.7\%), key trajectory features retained their sign and statistical significance. Specifically, global coherence remained negatively associated with creativity (LLM: $\beta = -0.400$ vs. $-0.211$, both $p < 0.001$), while semantic spread remained positively associated (LLM: $\beta = 0.227$ vs. $0.131$, both $p < 0.001$). This consistency across embedding models with different capacities and training procedures supports the validity of our trajectory-based approach.

\subsubsection{Creativity Operationalization Robustness} Our main analysis defined creativity as the product of novelty and usefulness (Novelty $\times$ Usefulness), following established practice in creativity research. To test sensitivity to this choice, we re-ran all analyses using an additive operationalization (Novelty + Usefulness). Table~\ref{tab:robustness_additive} demonstrates highly consistent results: coefficient patterns were nearly identical, and the core LLM creativity advantage remained robust (multiplicative: Cohen's $d = 1.50$; additive: $d = 1.37$, both $p < 0.001$). The slight decrease in effect size for the additive measure is expected, as the multiplicative definition better captures the synergistic relationship between novelty and usefulness that defines genuinely creative ideas.

\subsubsection{Inference-Time Scaling Control}
To rule out the possibility that discussion structure effects reflect inference-time scaling (i.e., more tokens leading to higher creativity), we regressed creativity on discussion structure while controlling for total token count. As shown in Table~\ref{tab:robustness_tokens}, token count contributed negligible additional variance ($\Delta R^2 = 0.06\%$) and had a non-significant coefficient near zero. Discussion structure effects (iterative vs. progressive: $\beta=+0.016$, $p<0.01$; open vs. progressive: $\beta=-0.030$, $p<0.001$) remained significant with this control, indicating that creativity gains reflect discussion structure rather than computational scaling.

\section{Supporting Figures}

\clearpage
\section{Supporting Tables}

\begin{table}[h]
\centering
\caption{Interclass Correlation Coefficients and 95\% Confidence Intervals for Novelty and Usefulness Ratings by Task}
\label{tab:ICC}
\begin{tabular}{llrrrr}
\toprule
\textbf{Task} & \textbf{Dimension} & \textbf{N} & \textbf{ICC} & \textbf{Lower Bound} & \textbf{Upper Bound} \\
\midrule
Education Inequality & Novelty & 855 & 0.87 & 0.86 & 0.88 \\
 & Usefulness & 855 & 0.80 & 0.78 & 0.82 \\
Employee Attrition & Novelty & 855 & 0.73 & 0.70 & 0.75 \\
 & Usefulness & 855 & 0.74 & 0.71 & 0.77 \\
Plastic Waste & Novelty & 839 & 0.78 & 0.76 & 0.80 \\
 & Usefulness & 839 & 0.70 & 0.67 & 0.73 \\
Singing in Shower & Novelty & 855 & 0.83 & 0.81 & 0.85 \\
 & Usefulness & 855 & 0.73 & 0.70 & 0.75 \\
Sorry Pandemic & Novelty & 839 & 0.89 & 0.88 & 0.90 \\
 & Usefulness & 839 & 0.79 & 0.77 & 0.82 \\
Supply Chain & Novelty & 839 & 0.90 & 0.89 & 0.91 \\
 & Usefulness & 839 & 0.76 & 0.73 & 0.79 \\
\bottomrule
\end{tabular}
\begin{tablenotes}
\small
\item Note: All Values are $ICC_{3,5}$, two-way mixed-effects average measures of five raters. N represents the number of ideas evaluated.
\end{tablenotes}
\end{table}

\begin{table}[h]
\centering
\caption{Descriptive Statistics for Creativity Ratings by Task and Group}
\label{tab:descriptive_stats_overall}
\begin{tabular}{llrrrr}
\toprule
\textbf{Task} & \textbf{Group} & \textbf{N} & \textbf{Creativity} & \textbf{Novelty} & \textbf{Usefulness} \\
\midrule
\textbf{Overall} & Human & 341 & 0.15 (0.09) & 0.31 (0.21) & 0.57 (0.21) \\
 & LLM & 4720 & 0.30 (0.10) & 0.53 (0.17) & 0.59 (0.16) \\
\midrule
Education Inequality & Human & 65 & 0.18 (0.09) & 0.32 (0.17) & 0.62 (0.20) \\
 & LLM & 788 & 0.34 (0.08) & 0.52 (0.14) & 0.68 (0.14) \\
Employee Attrition & Human & 65 & 0.18 (0.11) & 0.36 (0.20) & 0.53 (0.19) \\
 & LLM & 787 & 0.28 (0.10) & 0.49 (0.13) & 0.56 (0.10) \\
Plastic Waste & Human & 49 & 0.13 (0.09) & 0.31 (0.25) & 0.56 (0.26) \\
 & LLM & 790 & 0.32 (0.13) & 0.54 (0.20) & 0.60 (0.13) \\
Singing in Shower & Human & 64 & 0.12 (0.08) & 0.21 (0.13) & 0.57 (0.16) \\
 & LLM & 784 & 0.24 (0.08) & 0.54 (0.18) & 0.47 (0.16) \\
Sorry Pandemic & Human & 49 & 0.16 (0.08) & 0.39 (0.25) & 0.56 (0.27) \\
 & LLM & 781 & 0.28 (0.07) & 0.55 (0.18) & 0.56 (0.19) \\
Supply Chain & Human & 49 & 0.13 (0.07) & 0.27 (0.24) & 0.59 (0.21) \\
 & LLM & 790 & 0.33 (0.08) & 0.52 (0.17) & 0.66 (0.15) \\
\bottomrule
\end{tabular}
\begin{tablenotes}
\small
\item Note: Values shown as Mean (SD). All ratings were min-max scaled to the range [0, 1] within each task and rating dimension. LLM counts include 179 single-agent baseline ideas; the main human--LLM comparison uses multi-agent ideas only ($N=4{,}541$). N represents the number of ideas evaluated.
\end{tablenotes}
\end{table}
\FloatBarrier

\begingroup
\small
\setlength{\tabcolsep}{4pt} %

\begin{longtable}{@{}cccccccc@{}}
\caption{Complete experimental design matrix showing all 71 conditions tested. GPT-4.1 was run on all conditions; o3-high, o3-low, and mixed models were run only on three-agent ``different'' persona conditions.} \label{tab:conditions} \\
\toprule
\textbf{Cond.} & \textbf{Agents} & \textbf{Persona} & \textbf{Generation} & \textbf{Discussion} & \textbf{Pool} & \textbf{Length} & \textbf{Order} \\
\midrule
\endfirsthead

\multicolumn{8}{c}%
{{\tablename\ \thetable{} -- continued from previous page}} \\
\toprule
\textbf{Cond.} & \textbf{Agents} & \textbf{Persona} & \textbf{Generation} & \textbf{Discussion} & \textbf{Pool} & \textbf{Length} & \textbf{Order} \\
\midrule
\endhead

\midrule
\multicolumn{8}{r}{{Continued on next page}} \\
\endfoot

\bottomrule
\endlastfoot

1 & 3 & none & interactive & - & - & - & - \\
2 & 3 & same & interactive & - & - & - & - \\
3 & 3 & different & interactive & - & - & - & - \\
4 & 6 & none & interactive & - & - & - & - \\
5 & 6 & same & interactive & - & - & - & - \\
6 & 6 & different & interactive & - & - & - & - \\
7 & 3 & different & nominal & - & - & - & - \\
8 & 6 & different & nominal & - & - & - & - \\
\midrule
9 & 3 & none & - & open & - & 30 & fix \\
10 & 3 & same & - & open & - & 30 & fix \\
11 & 3 & different & - & open & - & 30 & fix \\
12 & 3 & different & - & open & - & 30 & random \\
13 & 3 & none & - & open & - & 60 & fix \\
14 & 3 & same & - & open & - & 60 & fix \\
15 & 3 & different & - & open & - & 60 & fix \\
16 & 3 & different & - & open & - & 60 & random \\
\midrule
17 & 3 & none & - & instructed & no & 30 & fix \\
18 & 3 & same & - & instructed & no & 30 & fix \\
19 & 3 & different & - & instructed & no & 30 & fix \\
20 & 3 & different & - & instructed & no & 30 & raise \\
21 & 3 & different & - & instructed & no & 30 & random \\
22 & 3 & none & - & instructed & no & 60 (min 30) & fix \\
23 & 3 & same & - & instructed & no & 60 (min 30) & fix \\
24 & 3 & different & - & instructed & no & 60 (min 30) & fix \\
25 & 3 & different & - & instructed & no & 60 (min 30) & raise \\
26 & 3 & different & - & instructed & no & 60 (min 30) & random \\
27 & 3 & none & - & iterative & - & 30 & fix \\
28 & 3 & same & - & iterative & - & 30 & fix \\
29 & 3 & different & - & iterative & - & 30 & fix \\
30 & 3 & different & - & iterative & - & 30 & random \\
\midrule
31 & 3 & none & interactive & instructed & no & 30 & fix \\
32 & 3 & same & interactive & instructed & no & 30 & fix \\
33 & 3 & different & interactive & instructed & no & 30 & fix \\
34 & 3 & different & interactive & instructed & no & 30 & raise \\
35 & 3 & different & interactive & instructed & no & 30 & random \\
36 & 3 & none & interactive & instructed & no & 60 (min 30) & fix \\
37 & 3 & same & interactive & instructed & no & 60 (min 30) & fix \\
38 & 3 & different & interactive & instructed & no & 60 (min 30) & fix \\
39 & 3 & different & interactive & instructed & no & 60 (min 30) & raise \\
40 & 3 & different & interactive & instructed & no & 60 (min 30) & random \\
41 & 3 & none & interactive & instructed & top5 & 30 & fix \\
42 & 3 & same & interactive & instructed & top5 & 30 & fix \\
43 & 3 & different & interactive & instructed & top5 & 30 & fix \\
44 & 3 & different & interactive & instructed & top5 & 30 & raise \\
45 & 3 & different & interactive & instructed & top5 & 30 & random \\
46 & 6 & none & interactive & instructed & no & 30 & fix \\
47 & 6 & same & interactive & instructed & no & 30 & fix \\
48 & 6 & different & interactive & instructed & no & 30 & fix \\
49 & 6 & different & interactive & instructed & no & 30 & raise \\
50 & 6 & different & interactive & instructed & no & 30 & random \\
51 & 6 & none & interactive & instructed & no & 60 (min 30) & fix \\
52 & 6 & same & interactive & instructed & no & 60 (min 30) & fix \\
53 & 6 & different & interactive & instructed & no & 60 (min 30) & fix \\
54 & 6 & different & interactive & instructed & no & 60 (min 30) & raise \\
55 & 6 & different & interactive & instructed & no & 60 (min 30) & random \\
56 & 3 & none & interactive & iterative & - & 30 & fix \\
57 & 3 & same & interactive & iterative & - & 30 & fix \\
58 & 3 & different & interactive & iterative & - & 30 & fix \\
59 & 3 & different & interactive & iterative & - & 30 & random \\
60 & 6 & none & interactive & iterative & - & 30 & fix \\
61 & 6 & same & interactive & iterative & - & 30 & fix \\
62 & 6 & different & interactive & iterative & - & 30 & fix \\
63 & 6 & different & interactive & iterative & - & 30 & random \\
\midrule
64 & 3 & none & interactive & progressive & - & - & fix \\
65 & 3 & same & interactive & progressive & - & - & fix \\
66 & 3 & different & interactive & progressive & - & - & fix \\
67 & 3 & different & interactive & progressive & - & - & random \\
68 & 6 & none & interactive & progressive & - & - & fix \\
69 & 6 & same & interactive & progressive & - & - & fix \\
70 & 6 & different & interactive & progressive & - & - & fix \\
71 & 6 & different & interactive & progressive & - & - & random \\
\end{longtable}
\endgroup

\begin{table*}[!htbp]
\tiny
\centering
\caption{\textbf{Hierarchical regression models predicting creativity from discussion structure and controls.}
OLS regression results with robust standard errors (HC3). Model 1 includes discussion structure only; Models 2--4 sequentially add model type, task fixed effects, and the model $\times$ discussion interaction. Model 4 is the primary specification. Data are restricted to three-agent teams with ``different'' persona assignments (N = 3,223 ideas from 26 conditions $\times$ 4 models). Reference categories: discussion = none, model = Mixed, task = education inequality. Standard errors are shown in parentheses. $^{*}p<0.05$, $^{**}p<0.01$, $^{***}p<0.001$.}
\label{tab:regression_full}

\setlength{\tabcolsep}{3pt}
\renewcommand{\arraystretch}{1.05}

\begin{tabular}{@{}lcccc@{}}
\hline
 & (1) & (2) & (3) & (4) \\
 & Discuss & +Model & +Task & +Interact \\
\hline

Progressive discussion
& 0.0207** & 0.0206** & 0.0205** & -0.0047 \\
& (0.0097) & (0.0094) & (0.0087) & (0.0172) \\

Progressive $\times$ GPT-4.1
& & & & 0.0884*** \\
& & & & (0.0226) \\

Progressive $\times$ o3-high
& & & & 0.0071 \\
& & & & (0.0250) \\

Progressive $\times$ o3-low
& & & & 0.0053 \\
& & & & (0.0249) \\

Instructed discussion
& 0.0202*** & 0.0202*** & 0.0200*** & -0.0192 \\
& (0.0075) & (0.0070) & (0.0067) & (0.0132) \\

Instructed $\times$ GPT-4.1
& & & & 0.0845*** \\
& & & & (0.0170) \\

Instructed $\times$ o3-high
& & & & 0.0427** \\
& & & & (0.0194) \\

Instructed $\times$ o3-low
& & & & 0.0296 \\
& & & & (0.0189) \\

Iterative discussion
& 0.0318*** & 0.0318*** & 0.0317*** & -0.0049 \\
& (0.0083) & (0.0079) & (0.0075) & (0.0146) \\

Iterative $\times$ GPT-4.1
& & & & 0.1015*** \\
& & & & (0.0195) \\

Iterative $\times$ o3-high
& & & & 0.0261 \\
& & & & (0.0214) \\

Iterative $\times$ o3-low
& & & & 0.0187 \\
& & & & (0.0212) \\

Open discussion
& -0.0083 & -0.0083 & -0.0083 & -0.0301** \\
& (0.0082) & (0.0076) & (0.0074) & (0.0142) \\

Open $\times$ GPT-4.1
& & & & 0.0429** \\
& & & & (0.0186) \\

Open $\times$ o3-high
& & & & 0.0380* \\
& & & & (0.0215) \\

Open $\times$ o3-low
& & & & 0.0063 \\
& & & & (0.0209) \\

GPT-4.1
& & -0.0272*** & -0.0267*** & -0.1015*** \\
& & (0.0046) & (0.0045) & (0.0159) \\

o3-high
& & 0.0417*** & 0.0422*** & 0.0086 \\
& & (0.0050) & (0.0046) & (0.0184) \\

o3-low
& & 0.0378*** & 0.0384*** & 0.0180 \\
& & (0.0049) & (0.0046) & (0.0179) \\

Employee attrition
& & & -0.0621*** & -0.0621*** \\
& & & (0.0053) & (0.0053) \\

Plastic waste
& & & -0.0145** & -0.0145** \\
& & & (0.0061) & (0.0061) \\

Singing in shower
& & & -0.1018*** & -0.1019*** \\
& & & (0.0053) & (0.0052) \\

Sorry pandemic
& & & -0.0661*** & -0.0662*** \\
& & & (0.0047) & (0.0047) \\

Supply chain
& & & -0.0127*** & -0.0127*** \\
& & & (0.0047) & (0.0046) \\

Intercept
& 0.2943*** & 0.2812*** & 0.3237*** & 0.3559*** \\
& (0.0071) & (0.0073) & (0.0076) & (0.0129) \\

\hline
$R^2$       & 0.015 & 0.096 & 0.227 & 0.241 \\
Adj. $R^2$  & 0.014 & 0.094 & 0.224 & 0.235 \\
N           & 3223  & 3223  & 3223  & 3223 \\
\hline
\end{tabular}
\end{table*}

\begin{table}
\caption{Discussion structure effects on creativity by model type (vs. no discussion baseline). Values show estimated effects from Model 4 with model $\times$ discussion interaction. For Mixed model, coefficients represent main effects; for other models, coefficients represent total effects (main + interaction). Standard errors in parentheses. $^{*}p<0.05$, $^{**}p<0.01$, $^{***}p<0.001$.}
\label{tab:discussion_effects}
\begin{tabular}{lllll}
\toprule
Model & Mixed & GPT-4.1 & o3-high & o3-low \\
Discussion &  &  &  &  \\
\midrule
Progressive & -0.0047 & 0.0838*** & 0.0024 & 0.0006 \\
Instructed & -0.0192 & 0.0653*** & 0.0234* & 0.0104 \\
Iterative & -0.0049 & 0.0966*** & 0.0212 & 0.0138 \\
Open & -0.0301* & 0.0128* & 0.0079 & -0.0238 \\
\bottomrule
\end{tabular}
\end{table}

\begin{table}
\caption{Discussion order effects within discussion conditions. OLS regression testing whether discussion order (fixed rotation vs hand raising vs random selection) affects creativity, controlling for discussion structure, model, and task. Left column: all conditions with discussion (N=2,983 from three-agent teams with different personas). Right column: instructed discussion only, where all models have all three order types (N=1,785). Reference category: random order. Standard errors in parentheses. $^{*}p<0.05$, $^{**}p<0.01$, $^{***}p<0.001$.}
\label{tab:discussion_order_effects}
\begin{tabular}{lcc}
\toprule
 & All Discussion & Instructed Only \\
 & (N=2983) & (N=1785) \\
\midrule
Fix vs Random & -0.0030 & -0.0007 \\
 & (0.0035) & (0.0050) \\
 & p=0.404 & p=0.894 \\
Raise vs Random & -0.0148** & -0.0137** \\
 & (0.0049) & (0.0052) \\
 & p=0.003 & p=0.009 \\
\midrule
R² & 0.230 & 0.254 \\
\bottomrule
\end{tabular}
\end{table}

\begin{table}
\caption{Discussion length effects. OLS regression testing whether discussion length (30 fixed rounds vs 60 fixed rounds vs 60 rounds with minimum 30) affects creativity, controlling for discussion structure, model, persona, team size, and task. Reference category: 30 rounds. Sample: N=3,702 observations from fixed-round discussion protocols (excludes progressive and none). Standard errors in parentheses.}
\label{tab:round_effects}
\begin{tabular}{lc}
\toprule
 & Creativity \\
\midrule
60 rounds vs 30 & -0.0056 \\
 & (0.0065) \\
 & p=0.392 \\
60\_min30 vs 30 & 0.0002 \\
 & (0.0036) \\
 & p=0.958 \\
\midrule
Omnibus test (LR) & p=0.719 \\
R² & 0.240 \\
$\Delta$R² (vs no rounds) & 0.000 \\
\bottomrule
\end{tabular}
\end{table}

\begin{table}
\caption{Idea generation mode effects in generation-only conditions (no discussion). OLS regression testing whether nominal (independent) vs interactive generation affects creativity, controlling for model, team size, and task. Sample: N=240 ideas from three-agent teams with different personas (conditions 3 and 7 from Table~\ref{tab:conditions}). Reference: interactive generation. Standard errors in parentheses. $^{*}p<0.05$, $^{**}p<0.01$, $^{***}p<0.001$.}
\label{tab:generation_effects}
\begin{tabular}{lc}
\toprule
 & Creativity \\
\midrule
Nominal vs Interactive & -0.0327** \\
 & (0.0123) \\
 & p=0.008 \\
\midrule
R² & 0.301 \\
\bottomrule
\end{tabular}
\end{table}

\begin{table}
\caption{Descriptive statistics for creativity, novelty, and usefulness by model type and discussion structure. Values show mean (standard error). Data restricted to conditions common across all models.}
\label{tab:descriptive_stats_discussion_methods}
\begin{tabular}{llrrrr}
\toprule
Model & Discussion & N & Creativity & Novelty & Usefulness \\
\midrule
GPT-4.1 & None & 60 & 0.212 (0.011) & 0.313 (0.014) & 0.674 (0.020) \\
GPT-4.1 & Progressive & 59 & 0.296 (0.012) & 0.483 (0.016) & 0.620 (0.020) \\
GPT-4.1 & Instructed & 450 & 0.277 (0.004) & 0.486 (0.007) & 0.586 (0.007) \\
GPT-4.1 & Iterative & 120 & 0.308 (0.008) & 0.613 (0.015) & 0.528 (0.015) \\
GPT-4.1 & Open & 120 & 0.224 (0.007) & 0.342 (0.009) & 0.658 (0.013) \\
o3-high & None & 60 & 0.322 (0.014) & 0.506 (0.017) & 0.646 (0.020) \\
o3-high & Progressive & 60 & 0.324 (0.013) & 0.673 (0.016) & 0.491 (0.019) \\
o3-high & Instructed & 450 & 0.345 (0.005) & 0.654 (0.007) & 0.544 (0.007) \\
o3-high & Iterative & 120 & 0.343 (0.009) & 0.577 (0.013) & 0.605 (0.012) \\
o3-high & Open & 120 & 0.329 (0.008) & 0.465 (0.010) & 0.711 (0.011) \\
o3-low & None & 60 & 0.331 (0.013) & 0.525 (0.018) & 0.644 (0.020) \\
o3-low & Progressive & 60 & 0.332 (0.014) & 0.663 (0.022) & 0.514 (0.020) \\
o3-low & Instructed & 448 & 0.341 (0.005) & 0.640 (0.007) & 0.547 (0.007) \\
o3-low & Iterative & 120 & 0.345 (0.009) & 0.596 (0.013) & 0.598 (0.015) \\
o3-low & Open & 120 & 0.307 (0.008) & 0.452 (0.011) & 0.687 (0.012) \\
Mixed & None & 60 & 0.313 (0.013) & 0.442 (0.014) & 0.713 (0.019) \\
Mixed & Progressive & 60 & 0.308 (0.013) & 0.555 (0.022) & 0.572 (0.019) \\
Mixed & Instructed & 438 & 0.295 (0.005) & 0.516 (0.007) & 0.584 (0.008) \\
Mixed & Iterative & 119 & 0.309 (0.008) & 0.533 (0.013) & 0.603 (0.016) \\
Mixed & Open & 120 & 0.283 (0.007) & 0.421 (0.009) & 0.675 (0.011) \\
\bottomrule
\end{tabular}
\end{table}

\begin{table}[h]
\centering
\caption{\textbf{Distribution of top-10 ideas per task by source.} 
For each task, we identified the top 10 ideas ranked by creativity, novelty, and usefulness separately, 
then counted how many came from human teams versus LLM teams.}
\label{tab:top10_distribution}
\begin{tabular}{l|cc|cc|cc}
\hline
 & \multicolumn{2}{c|}{\textbf{Creativity}} & \multicolumn{2}{c|}{\textbf{Novelty}} & \multicolumn{2}{c}{\textbf{Usefulness}} \\
\textbf{Task} & Human & LLM & Human & LLM & Human & LLM \\
\hline
Education Inequality & 0 & 10 & 1 & 9 & 3 & 7 \\
Sorry Pandemic & 0 & 10 & 3 & 7 & 4 & 6 \\
Supply Chain & 0 & 10 & 3 & 7 & 2 & 8 \\
Employee Attrition & 0 & 10 & 4 & 6 & 2 & 8 \\
Singing in Shower & 0 & 10 & 0 & 10 & 3 & 7 \\
Plastic Waste & 0 & 10 & 1 & 9 & 1 & 9 \\
Total & 0 & 60 & 12 & 48 & 15 & 45 \\
\hline
\end{tabular}
\end{table}

\FloatBarrier

\begin{longtable}{cp{9cm}c}
\caption{\textbf{Top 5 highest-scoring ideas per task.} Ideas are shown as harmonized paraphrases used for blinded evaluation. All top-scoring ideas were generated by LLM teams.}
\label{tab:top_ideas} \\
\hline
\textbf{\#} & \textbf{Idea} & \textbf{Creativity} \\
\hline
\endfirsthead
\hline
\textbf{\#} & \textbf{Idea} & \textbf{Creativity} \\
\hline
\endhead
\hline
\endfoot
\multicolumn{3}{l}{\textbf{Education Inequality}} \\
\hline
1 & Village paths are turned into interactive learning trails. Special ink shows lessons during the day, and glowing arrows guide practice at night. Raised symbols and Braille help blind people learn by touch. Walking on the path generates power for small radio beacons that play short lessons, which can be heard on any hand-crank radio. Learners answer quiz questions to earn small cash rewards and help fund school meals. Local teenagers are paid to repaint and update the lessons. No phones or internet are needed. & 0.62 \\
2 & A necklace made from recycled fishing line holds NFC beads that give access to stories and can also be used as a measuring tape to help with math. A hand-cranked scanner charges a small battery to power a light for reading at night. Blank beads come with stickers so teachers can easily label them in different languages. Repair stations let students fix or remake broken beads themselves, earning more beads for different subjects and making the necklace appealing for all ages. & 0.62 \\
3 & Develop a National School Equity Index that measures school funding and student success using up-to-date data on local taxes, student results, and living costs. If a district scores too low, it automatically gets extra money collected from a small fee on financial trades. This money goes straight to schools through secure digital accounts and can only be used for proven needs like hiring teachers, buying science kits, or adding counselors. The index also shows public rankings to encourage schools and donors to work harder at reducing differences between districts. & 0.62 \\
4 & Reality Reboot Pods are flexible learning spaces where students, especially those from underserved groups, can shape their entire learning environment. Each Pod includes a small living ecosystem, AI tools that provide information about local issues, and interactive digital interfaces. Students can create solutions that affect both their immediate surroundings and larger communities, such as cleaning water or testing ideas in virtual reality. The most effective student projects receive immediate support and guidance from a worldwide network. This approach makes learning and making real-world changes happen at the same time. & 0.62 \\
5 & This idea reuses plastic water jugs by turning their built-in lens into a simple projector. At night, children attach a cap with tiny printed lessons to a basic LED flashlight, which projects clear images onto a wall. Each cap contains different educational topics and costs very little. After discussing each image, kids click to the next one, and the cap secretly records their answers. A local kiosk reads these answers and gives rewards like phone data or meal credits. During the day, the jug can also be used to clean water using sunlight, combining education with health benefits. & 0.60 \\
\hline
\multicolumn{3}{l}{\textbf{Sorry Pandemic}} \\
\hline
1 & A system uses three steps to stop people from saying the word “sorry” out loud. First, a neck band senses when someone is about to say the word and gently vibrates to pause them, while also rewarding polite behavior. Second, people can show they’re sorry using a simple gesture or a quick sound or vibration on their device, which everyone understands. Third, keyboards and captions automatically replace “sorry” with a similar-sounding word. This approach helps people stay polite without spreading the word and can be used easily and affordably everywhere. & 0.62 \\
2 & A system helps people avoid spreading the word “sorry” by using three steps: a special toothpaste changes how the “s” sound comes out, phone and earbud software blocks the sound before it’s spoken and tracks avoided apologies, and a double-tap gesture on the chest silently shows regret and gives small rewards. These physical, digital, and reward methods work together so people can still apologize without spreading the word, helping reduce its use without strict rules. & 0.57 \\
3 & A simple double-tap over the heart becomes a universal way to say sorry. Devices like phones, watches, or wristbands sense the gesture, show a green heart, give a gentle buzz, and can send a small donation to local charities. If someone doesn’t have a device, the gesture alone still shows they’re sorry. Devices can also gently remind people before they say something hurtful. Public places use the same heart symbol, making it easy for everyone to apologize quietly and clearly without needing to say the word "sorry." & 0.53 \\
4 & A pack of ten temporary tattoos uses special ink to sense when someone is about to say “sorry.” Worn near the jaw, each tattoo connects wirelessly to a smartphone. When it detects the start of the word, the phone vibrates and flashes, and the tattoo gives a gentle buzz to stop the person from speaking. Instead, they are encouraged to use a silent hand-over-heart gesture. The phone collects anonymous data to help public health efforts. The tattoos are cheap, single-use, and designed for easy, everyday use. & 0.50 \\
5 & This is a minty, edible strip that sticks to the roof of the mouth. Tiny sensors in the strip detect when someone starts to say “sorry” and quickly close to block breath and germs from escaping. The sensors open for all other sounds, so talking feels normal. The strip has color spots that fade each time it blocks a “sorry,” reminding people to use a silent, hand-over-heart gesture instead. Each strip is cheap, biodegradable, and sold in vending machines, making it an easy and effective way to prevent spreading germs. & 0.50 \\
\hline
\multicolumn{3}{l}{\textbf{Supply Chain}} \\
\hline
1 & The Adaptive Immune Supply Chain Matrix is a system where each supply point works like an immune cell, detecting problems with advanced sensors and quickly responding by creating digital defenses such as smart contracts or decoy shipments. The system learns from both real and simulated disruptions, improving its ability to handle future issues by updating its processes. Human managers design new ways to protect the supply chain using interactive data tools, making the entire network smarter and better at preventing and responding to risks and extra costs. & 0.58 \\
2 & A thin, edible tag made from rice paper contains tiny compartments with dried bacteria spores, each sealed with a natural film. When a package is scanned with a special light, the tag creates a colorful QR code that records the time and tracks the package’s journey. If the package gets too hot or is exposed to chemicals, the tag changes color to show it’s spoiled. Any smartphone can read the code and automatically record the information. The tag is safe to eat, compostable, contains no metal, and is very cheap to make. & 0.55 \\
3 & This idea describes a pallet wrap made from two clear plastic layers with a special gel in between that contains tiny, breakable crystals. At each checkpoint, a magnetic pulse breaks some of these crystals, changing the color pattern in the wrap. A phone scans this pattern and records it in a digital ledger. If someone tampers with the wrap, the pattern changes in a way that shows it was opened. When the shipment arrives, staff wash away the gel, leaving only the recyclable plastic. & 0.53 \\
4 & Create a supply chain system that uses smart robots and digital traps to spot and respond to threats. The system sends out fake shipments and digital decoys that act like real goods, tricking attackers and collecting information. These decoys can change their routes or pretend to fail, sharing real-time data with a network of artificial intelligence. The AI quickly adapts, closing security gaps and improving delivery routes. This makes the supply chain stronger and harder to predict, learning from each attack to better protect itself in the future. & 0.53 \\
5 & A product is shipped in an unfinished state. The final step to make it work depends on special instructions created by AI using data collected from the product’s journey, like movement and light exposure. At the destination, this data is used to complete the product. If someone steals the unfinished product, they can’t use it because the instructions needed to finish it are unique to its original journey. & 0.53 \\
\hline
\multicolumn{3}{l}{\textbf{Employee Attrition}} \\
\hline
1 & Create a program where employees can pitch any big idea, even if it’s not related to the company’s work. Chosen ideas get funding, time off to work on them, and advice from leaders and outside experts. If the idea succeeds, employees can either start their own business or bring it back as a new part of the company, leading its growth. Employees keep their jobs and career options no matter what happens, so they can take risks without worry. This makes the company a place where people feel free to pursue their biggest goals. & 0.65 \\
2 & Every three months, each employee gets 100 digital shares tracked on a secure system. Coworkers, managers, former staff, and some clients can use extra resources or time to bid for these shares. An AI updates share values weekly based on positive actions like learning new skills or helping others, and lowers them if signs of stress appear. If a share’s value drops by 15\%, everyone who owns it must offer real help within three days or lose value. Employees earn flexible rewards based on their share value each payday. If someone quits, their shares fund quick projects to improve retention. & 0.58 \\
3 & When an employee creates a major improvement, they get a permanent right to use 20\% of their work time however they choose. This time is protected and funded, letting them manage their own projects, help others, learn new skills, or take short breaks. This reward is ongoing, giving them lasting freedom and trust instead of just a one-time benefit. & 0.58 \\
4 & Each employee can choose one important person in their life, like a spouse or close friend, to get free access to the company’s training, career coaching, and short-term job opportunities. Any money or certificates this person earns help increase the employee’s bonuses and paid time off. By supporting someone the employee cares about, the company helps keep employees from leaving for better opportunities elsewhere and builds stronger loyalty. & 0.57 \\
5 & Every three months, employees set three personal and team goals for the next year. They create 100 digital tokens linked to these goals. Others can invest in these tokens, but only if they help the employee in some way, like offering advice or making a connection. Tokens become active when at least three people pledge support. The system tracks which employees might leave. After a year, any rewards are shared between investors, the employee (as learning credits), and supporters. If goals aren’t met, the tokens fund referrals and training. Leaders’ bonuses depend on how many tokens are active. & 0.57 \\
\hline
\multicolumn{3}{l}{\textbf{Singing in Shower}} \\
\hline
1 & Create a language-learning tool that turns vocabulary words into short, easy-to-sing songs for people to listen to while showering. Users pick the language they want to learn, and the service makes songs using their target words, which play through waterproof speakers. The system listens for correct pronunciation and gives instant feedback, unlocking new words as users improve. This approach uses repetition and the relaxing shower environment to help people remember words better and enjoy learning about new cultures. & 0.77 \\
2 & Employees can earn extra paid time off by singing in the shower for 20 minutes. A small waterproof sticker works with a smartwatch to confirm real singing without recording any sound, protecting privacy. After five sessions, employees get an extra hour of flexible leave, which they can track online. This encourages people to start their day in a fun way, helps them feel better, and gives companies a new way to support employee well-being. & 0.56 \\
3 & AquaVerse Odyssey is an interactive adventure where people’s shower singing sessions create unique characters with their own skills and personalities. Players join global teams to complete story-based missions, like restoring digital rivers and solving environmental challenges. Team progress depends on group mood, sharing resources, and real-world actions such as organizing water cleanups or fundraising. Reaching big goals unlocks virtual water festivals, health tips, and documentaries showing the group’s impact, turning singing into a fun way to help the environment and make a real difference. & 0.55 \\
4 & A smart shower uses special waterproof tiles and artificial intelligence to change the look, smell, and sound of the shower based on how someone sings. As singing continues, the tiles show moving images, release matching scents, and play sounds like rain or waterfalls that fit the mood of the song. The longer and more passionately someone sings, the more intense and immersive the experience becomes. After each session, the system creates a unique digital record of the sights, sounds, and scents from the shower, making each experience personal and memorable. & 0.53 \\
5 & A small, water-powered device sticks to the shower wall and projects karaoke lyrics onto the glass. It tracks how long someone sings, and after 20 minutes, it gives three rewards: adds money to a savings account, donates clean water (more if the water was turned off during singing), and creates a short remix of the person’s singing to share. Sharing the remix lets friends join in, encourages them to get the app, and leads to more water donations, making the process repeat itself. & 0.52 \\
\hline
\multicolumn{3}{l}{\textbf{Plastic Waste}} \\
\hline
1 & A program turns mixed plastic waste into strong pavement tiles that generate electricity when people walk on them. Local centers shred the plastic and make tiles with built-in devices that create power from footsteps. These tiles are used in busy public places like sidewalks and playgrounds. The electricity they produce helps power local services and provides free public Wi-Fi. An app shows how much energy each tile makes and how much plastic is reused. Old tiles are collected, recycled, and made into new ones, creating a system that reduces plastic waste and supplies clean energy. & 0.74 \\
2 & Wind-powered boats collect ocean plastic, break it down, and turn it into a material that feeds special algae. Robots on the boats use this mixture, along with crushed seashells and helpful bacteria, to 3D print reef tiles. Divers attach these tiles to damaged reefs, where seawater activates microbes that turn the tiles into limestone and encourage coral to grow. For every kilogram of plastic collected, one square meter of new reef is created, and fishers who help earn rewards. & 0.64 \\
3 & Biodegradable mesh barriers made from invasive plants are placed in rivers to catch plastic waste without blocking water or harming animals. Local groups collect the larger plastic pieces for recycling. The mesh contains safe fungi that break down tiny plastic particles as the mesh slowly dissolves. The leftover material is used to help native plants grow along the riverbanks. This system uses natural water flow to trap plastic, relies on fungi to remove microplastics, and involves the community in both cleanup and restoration, creating jobs and improving the river environment. & 0.64 \\
4 & Solar-powered helium blimps would fly over cities, releasing a gentle mist that pulls in tiny plastic and soot particles from the air. Built-in sensors track things like pollen and pollution, sending real-time updates to health officials so they can warn schools and hospitals if needed. When the blimps’ filters are full, they turn the collected pollution into safe, fire-resistant building materials for affordable housing. The blimps return to local stations for upkeep, creating new jobs and turning air pollution into useful products for the community. & 0.64 \\
5 & Place sprayers at river outlets that release a thin, clear, biodegradable film on the water during slow-flow times. This film attracts and gathers microplastics into clumps, which then float to collection points for easy removal. After twelve hours, the film safely breaks down into nutrients, so it doesn’t harm wildlife. Sensors track how well the system works and adjust spraying as needed, making it an energy-efficient and safe way to stop microplastics from reaching the ocean. & 0.64 \\
\hline
\end{longtable}

\begin{table}
\tiny
\centering
\caption{\textbf{Novelty-usefulness decomposition of discussion structure effects.} 
Regression coefficients from parallel models predicting creativity (Model 4 from main analysis), novelty, and usefulness separately. Discussion structures that boost novelty systematically reduce usefulness, revealing a fundamental trade-off. 
GPT-4.1's iterative discussion overcomes this trade-off by substantially increasing novelty ($\beta=+0.208$, $p<0.001$) without proportionally reducing usefulness. 
Reference categories: discussion = none, model = Mixed. 
Standard errors in parentheses. 
$^{*}p<0.05$, $^{**}p<0.01$, $^{***}p<0.001$.}
\label{tab:novelty_usefulness}
\begin{tabular}{llll}
\hline
                                                                   & Creativity & Novelty    & Usefulness  \\
\hline
C(discussion, Treatment("none"))[T.progressive]                       & -0.0061    & 0.0644***  & -0.0843***  \\
                                                                   & (0.0122)   & (0.0199)   & (0.0174)    \\
C(discussion, Treatment("none"))[T.progressive]:C(model)[T.gpt-4.1]   & 0.0885***  & 0.0572*    & 0.0858**    \\
                                                                   & (0.0229)   & (0.0337)   & (0.0343)    \\
C(discussion, Treatment("none"))[T.progressive]:C(model)[T.o3-high]   & 0.0071     & 0.0543     & -0.0148     \\
                                                                   & (0.0248)   & (0.0339)   & (0.0351)    \\
C(discussion, Treatment("none"))[T.progressive]:C(model)[T.o3-low]    & 0.0053     & 0.0257     & 0.0105      \\
                                                                   & (0.0248)   & (0.0384)   & (0.0369)    \\
C(discussion, Treatment("none"))[T.instructed]                     & -0.0159**  & 0.0266**   & -0.0661***  \\
                                                                   & (0.0078)   & (0.0104)   & (0.0117)    \\
C(discussion, Treatment("none"))[T.instructed]:C(model)[T.gpt-4.1] & 0.0847***  & 0.0984***  & 0.0425*     \\
                                                                   & (0.0173)   & (0.0231)   & (0.0257)    \\
C(discussion, Treatment("none"))[T.instructed]:C(model)[T.o3-high] & 0.0428**   & 0.0735***  & 0.0278      \\
                                                                   & (0.0192)   & (0.0236)   & (0.0272)    \\
C(discussion, Treatment("none"))[T.instructed]:C(model)[T.o3-low]  & 0.0298     & 0.0410*    & 0.0338      \\
                                                                   & (0.0189)   & (0.0248)   & (0.0278)    \\
C(discussion, Treatment("none"))[T.iterative]                      & -0.0058    & 0.0436***  & -0.0531***  \\
                                                                   & (0.0095)   & (0.0135)   & (0.0155)    \\
C(discussion, Treatment("none"))[T.iterative]:C(model)[T.gpt-4.1]  & 0.1015***  & 0.2082***  & -0.0359     \\
                                                                   & (0.0198)   & (0.0284)   & (0.0309)    \\
C(discussion, Treatment("none"))[T.iterative]:C(model)[T.o3-high]  & 0.0261     & -0.0206    & 0.0693**    \\
                                                                   & (0.0212)   & (0.0279)   & (0.0309)    \\
C(discussion, Treatment("none"))[T.iterative]:C(model)[T.o3-low]   & 0.0187     & -0.0202    & 0.0644**    \\
                                                                   & (0.0212)   & (0.0290)   & (0.0325)    \\
C(discussion, Treatment("none"))[T.open]                           & -0.0305*** & -0.0684*** & 0.0208      \\
                                                                   & (0.0094)   & (0.0123)   & (0.0137)    \\
C(discussion, Treatment("none"))[T.open]:C(model)[T.gpt-4.1]       & 0.0429**   & 0.0498**   & 0.0214      \\
                                                                   & (0.0189)   & (0.0248)   & (0.0281)    \\
C(discussion, Treatment("none"))[T.open]:C(model)[T.o3-high]       & 0.0380*    & -0.0191    & 0.1028***   \\
                                                                   & (0.0213)   & (0.0255)   & (0.0293)    \\
C(discussion, Treatment("none"))[T.open]:C(model)[T.o3-low]        & 0.0063     & -0.0515*   & 0.0807***   \\
                                                                   & (0.0208)   & (0.0272)   & (0.0298)    \\
C(discussion\_order\_plan, Treatment("not\_applicable"))[T.fix]    & -0.0165*   & 0.0263**   & -0.0595***  \\
                                                                   & (0.0089)   & (0.0111)   & (0.0135)    \\
C(discussion\_order\_plan, Treatment("not\_applicable"))[T.raise]  & -0.0283*** & 0.0219*    & -0.0752***  \\
                                                                   & (0.0095)   & (0.0122)   & (0.0144)    \\
C(discussion\_order\_plan, Treatment("not\_applicable"))[T.random] & -0.0135    & 0.0180     & -0.0479***  \\
                                                                   & (0.0089)   & (0.0110)   & (0.0135)    \\
C(idea\_generation\_plan, Treatment("none"))[T.interactive]        & 0.0011     & 0.0011     & 0.0026      \\
                                                                   & (0.0037)   & (0.0062)   & (0.0058)    \\
C(idea\_generation\_plan, Treatment("none"))[T.nominal]            & -0.0316**  & -0.0511*** & 0.0076      \\
                                                                   & (0.0127)   & (0.0168)   & (0.0187)    \\
C(model)[T.gpt-4.1]                                                & -0.1015*** & -0.1290*** & -0.0385     \\
                                                                   & (0.0162)   & (0.0208)   & (0.0240)    \\
C(model)[T.o3-high]                                                & 0.0086     & 0.0636***  & -0.0663***  \\
                                                                   & (0.0182)   & (0.0215)   & (0.0255)    \\
C(model)[T.o3-low]                                                 & 0.0180     & 0.0827***  & -0.0686***  \\
                                                                   & (0.0179)   & (0.0228)   & (0.0261)    \\
C(question\_id)[T.employee\_attrition]                             & -0.0621*** & -0.0309*** & -0.1061***  \\
                                                                   & (0.0053)   & (0.0073)   & (0.0074)    \\
C(question\_id)[T.plastic\_waste]                                  & -0.0145**  & 0.0317***  & -0.0741***  \\
                                                                   & (0.0061)   & (0.0082)   & (0.0074)    \\
C(question\_id)[T.singing\_shower]                                 & -0.1019*** & 0.0305***  & -0.2100***  \\
                                                                   & (0.0052)   & (0.0078)   & (0.0086)    \\
C(question\_id)[T.sorry\_pandemic]                                 & -0.0663*** & 0.0159*    & -0.1172***  \\
                                                                   & (0.0047)   & (0.0081)   & (0.0087)    \\
C(question\_id)[T.supply\_chain]                                   & -0.0127*** & 0.0085     & -0.0204**   \\
                                                                   & (0.0046)   & (0.0075)   & (0.0080)    \\
Intercept                                                          & 0.3712***  & 0.4580***  & 0.7957***   \\
                                                                   & (0.0155)   & (0.0187)   & (0.0233)    \\
R-squared                                                          & 0.2451     & 0.3208     & 0.3037      \\
R-squared Adj.                                                     & 0.2385     & 0.3148     & 0.2976      \\
Adj. R²                                                            & 0.239      & 0.315      & 0.298       \\
N                                                                  & 3223       & 3223       & 3223        \\
R²                                                                 & 0.245      & 0.321      & 0.304       \\
\hline
\end{tabular}
\end{table}

\begin{table}
\caption{Persona assignment effects on creativity, novelty, and usefulness (GPT-4.1 only, matched conditions). Values show mean (standard error). No significant differences were found in overall creativity (all pairwise comparisons p>0.05), but persona assignment shifted the relative balance between novelty and usefulness.}
\label{tab:persona_effects}
\begin{tabular}{lrrrr}
\toprule
Persona & N & Creativity & Novelty & Usefulness \\
\midrule
Same & 478 & 0.258 (0.004) & 0.461 (0.007) & 0.587 (0.007) \\
Different & 480 & 0.267 (0.004) & 0.480 (0.008) & 0.585 (0.007) \\
No & 480 & 0.269 (0.004) & 0.541 (0.008) & 0.534 (0.008) \\
\bottomrule
\end{tabular}
\end{table}

\begin{table}
\caption{Team size effects on creativity, novelty, and usefulness for LLM (GPT-4.1) and human teams. Values show mean (standard error). Both LLM and human teams exhibit process loss, with larger teams showing lower creativity. For GPT-4.1, data restricted to matched conditions with different persona assignments.}
\label{tab:team_size_llm_human}
\begin{tabular}{llrrrr}
\toprule
Source & Team Size & N & Creativity & Novelty & Usefulness \\
\midrule
GPT-4.1 & 3 agents & 359 & 0.280 (0.005) & 0.479 (0.008) & 0.606 (0.008) \\
GPT-4.1 & 6 agents & 359 & 0.271 (0.005) & 0.511 (0.009) & 0.558 (0.009) \\
Human & 3 agents & 224 & 0.155 (0.006) & 0.320 (0.014) & 0.553 (0.014) \\
Human & 6 agents & 117 & 0.142 (0.008) & 0.287 (0.019) & 0.600 (0.020) \\
\bottomrule
\end{tabular}
\end{table}

\begin{table}[h]
\centering
\caption{\textbf{Replacement pool effects on creativity.} 
Matched conditions differing only in replacement pool availability 
(conditions 41--45 vs.\ 31--35). Values show mean (standard error). 
$^{*}p<0.05$, $^{**}p<0.01$, $^{***}p<0.001$.}
\label{tab:replacement_pool}
\begin{tabular}{llcccc}
\toprule
\textbf{Model} & \textbf{Pool} & \textbf{N} & \textbf{Creativity} & \textbf{Novelty} & \textbf{Usefulness} \\
\midrule
GPT-4.1 & Pool & 150 & 0.275 (0.007) & 0.461 (0.009) & 0.606 (0.012) \\
GPT-4.1 & No pool & 149 & 0.275 (0.006) & 0.459 (0.011) & 0.612 (0.011) \\
\midrule
o3-high & Pool & 90 & 0.351 (0.011) & 0.659 (0.015) & 0.551 (0.018) \\
o3-high & No pool & 90 & 0.345 (0.011) & 0.642 (0.016) & 0.551 (0.016) \\
\midrule
o3-low & Pool & 90 & 0.343 (0.011) & 0.623 (0.014) & 0.565 (0.016) \\
o3-low & No pool & 90 & 0.346 (0.009) & 0.622 (0.015) & 0.572 (0.014) \\
\midrule
Mixed & Pool & 90 & 0.287 (0.011) & 0.477 (0.016) & 0.609 (0.018) \\
Mixed & No pool & 88 & 0.302 (0.011) & 0.538 (0.015) & 0.576 (0.018) \\
\midrule
\textbf{Overall} & Pool & 420 & 0.308 (0.005) & 0.541 (0.008) & 0.586 (0.008) \\
\textbf{Overall} & No pool & 417 & 0.311 (0.005) & 0.550 (0.008) & 0.583 (0.007) \\
 & $\Delta$ & & $d=-0.03$ & $d=-0.06$ & $d=0.02$ \\
\bottomrule
\end{tabular}
\end{table}

\begin{table}[!htbp]
\centering
\footnotesize
\caption{\textbf{Trajectory feature definitions.}
All features were computed from turn-level embeddings using Qwen3-Embedding-0.6B.
Here, $d_{\cos}$ denotes cosine distance, $e_i$ is the embedding at turn $i$,
$\bar e$ is the conversation centroid, and $n$ is the number of turns.}
\label{tab:trajectory_definitions}

\setlength{\tabcolsep}{3pt}
\renewcommand{\arraystretch}{1.25}

\begin{tabularx}{\textwidth}{@{}p{2.4cm}p{1.7cm}p{4.6cm}>{\raggedright\arraybackslash}X@{}}
\hline
\textbf{Feature} & \textbf{Category} & \textbf{Formula} & \textbf{Interpretation} \\
\hline

Local coherence
& Coherence
& $\frac{1}{n-1}\sum_{i=1}^{n-1}\left(1-d_{\cos}(e_i,e_{i+1})\right)$
& Turn-to-turn semantic continuity~\cite{iter2020automatic,elvevag2007quantifying}. \\

Global coherence
& Coherence
& $\frac{1}{n}\sum_{i=1}^{n}\left(1-d_{\cos}(e_i,\bar e)\right)$
& Focus around the conversation centroid~\cite{li2014model}. \\

Path length
& Exploration
& $\sum_{i=1}^{n-1} d_{\cos}(e_i,e_{i+1})$
& Total semantic movement across turns~\cite{malaie2024divergent,hills2015exploration}. \\

Convergence ratio
& Dynamics
& $(D_{\mathrm{early}}-D_{\mathrm{late}})/D_{\mathrm{early}}$, where $D$ is mean pairwise cosine distance
& Reduction in semantic dispersion from the first half to the second half of the conversation~\cite{guilford1967creativity,girotra2010idea}. \\

Max distance
& Exploration
& $\max_{i,j} d_{\cos}(e_i,e_j)$
& Maximum semantic separation between any two turns~\cite{johnson2023divergent,olson2021naming}. \\

Trajectory curvature
& Dynamics
& $\frac{1}{n-2}\sum_{i=1}^{n-2}
\arccos\left(
\frac{\Delta_i\cdot\Delta_{i+1}}
{\|\Delta_i\|_2\|\Delta_{i+1}\|_2}
\right)$, where $\Delta_i=e_{i+1}-e_i$
& Average angular change between consecutive semantic steps~\cite{hills2012optimal}. \\

Topic switching rate
& Dynamics
& Cluster transitions divided by $(n-1)$ using $k$-means with $k=3$
& Frequency of transitions between semantic clusters~\cite{troyer1997clustering,zarriess2025components}. \\

Revisit score
& Refinement
& $\max_{i\in L,\,j\in E}\left(1-d_{\cos}(e_i,e_j)\right)$, where $E$ and $L$ are the early and late 30\% of turns
& Degree to which late turns return to early semantic regions~\cite{coursey2019linking}. \\

Semantic spread
& Exploration
& $\mathrm{SD}\left(\|e_i-\bar e\|_2\right)$
& Dispersion of turns around the conversation centroid~\cite{kenett2014investigating}. \\

\hline
\end{tabularx}
\end{table}

\begin{table}
\tiny
\caption{\textbf{Hierarchical regression models predicting LLM creativity from trajectory features.} 
Models 1-4: OLS regression with robust standard errors (HC3). 
Model 5: Mixed effects with task random intercepts (REML estimation). 
Outcome variable: standardized creativity rating (z-scored within LLM sample). 
All trajectory features standardized (z-scored). 
Conditional R² includes both fixed and random effects; Marginal R² includes fixed effects only. 
Reference categories: Task = education inequality, Model = Mixed, Discussion = none. 
$^{*}p<0.10$, $^{**}p<0.05$, $^{***}p<0.01$.}
\label{tab:llm_regression_full}
\centering
\begin{tabular}{@{}lccccc@{}}
\hline
                                       & Model 1      & Model 2    & Model 3    & Model 4        & Model 5          \\
                                       & (Trajectory) & (+ Task)   & (+ Model)  & (+ Discussion) & (Mixed Effects)  \\
\hline
C(discussion)[T.iterative]             &              &            &            & 0.1365**       & 0.1381**         \\
                                       &              &            &            & (0.0566)       & (0.0566)         \\
C(discussion)[T.open]                  &              &            &            & -0.3425***     & -0.3425***       \\
                                       &              &            &            & (0.0473)       & (0.0473)         \\
C(models)[T.gpt-4.1]                   &              &            & -0.1943*** & -0.1910***     & -0.1896***       \\
                                       &              &            & (0.0515)   & (0.0511)       & (0.0511)         \\
C(models)[T.o3-high]                   &              &            & 0.4867***  & 0.4650***      & 0.4648***        \\
                                       &              &            & (0.0506)   & (0.0503)       & (0.0503)         \\
C(models)[T.o3-low]                    &              &            & 0.4019***  & 0.3891***      & 0.3890***        \\
                                       &              &            & (0.0515)   & (0.0511)       & (0.0511)         \\
C(question\_id)[T.employee\_attrition] &              & -0.5837*** & -0.5894*** & -0.5884***     &                  \\
                                       &              & (0.0527)   & (0.0512)   & (0.0508)       &                  \\
C(question\_id)[T.plastic\_waste]      &              & -0.3093*** & -0.2758*** & -0.2803***     &                  \\
                                       &              & (0.0527)   & (0.0513)   & (0.0508)       &                  \\
C(question\_id)[T.singing\_shower]     &              & -0.9261*** & -1.0037*** & -0.9804***     &                  \\
                                       &              & (0.0545)   & (0.0533)   & (0.0531)       &                  \\
C(question\_id)[T.sorry\_pandemic]     &              & -0.5816*** & -0.6032*** & -0.5870***     &                  \\
                                       &              & (0.0545)   & (0.0530)   & (0.0528)       &                  \\
C(question\_id)[T.supply\_chain]       &              & -0.1091**  & -0.0995*   & -0.0960*       &                  \\
                                       &              & (0.0528)   & (0.0513)   & (0.0508)       &                  \\
Group Var                              &              &            &            &                & 0.1731           \\
                                       &              &            &            &                & (0.1108)         \\
Intercept                              & -0.0000      & 0.4165***  & 0.3528***  & 0.3812***      & -0.0417          \\
                                       & (0.0160)     & (0.0377)   & (0.0505)   & (0.0534)       & (0.1542)         \\
R-squared                              & 0.0865       & 0.1761     & 0.2230     & 0.2378         &                  \\
R-squared Adj.                         & 0.0841       & 0.1728     & 0.2192     & 0.2337         &                  \\
convergence\_ratio                     & 0.0104       & -0.0226    & -0.0240    & -0.0040        & -0.0036          \\
                                       & (0.0181)     & (0.0173)   & (0.0168)   & (0.0169)       & (0.0169)         \\
global\_coherence                      & -0.5148***   & -0.4001*** & -0.1408*** & -0.1202***     & -0.1219***       \\
                                       & (0.0413)     & (0.0399)   & (0.0441)   & (0.0461)       & (0.0461)         \\
local\_coherence                       & -0.0164      & -0.0527    & 0.1021**   & 0.0186         & 0.0182           \\
                                       & (0.0472)     & (0.0451)   & (0.0453)   & (0.0469)       & (0.0469)         \\
max\_distance                          & -0.2698***   & -0.2758*** & -0.0792*** & -0.0493        & -0.0492          \\
                                       & (0.0292)     & (0.0278)   & (0.0306)   & (0.0362)       & (0.0362)         \\
path\_length                           & -0.1217***   & -0.1216*** & -0.0697*** & -0.0789***     & -0.0791***       \\
                                       & (0.0218)     & (0.0207)   & (0.0204)   & (0.0203)       & (0.0203)         \\
revisit\_score                         & -0.0026      & 0.0086     & 0.0089     & 0.0171         & 0.0169           \\
                                       & (0.0205)     & (0.0196)   & (0.0192)   & (0.0197)       & (0.0197)         \\
semantic\_spread                       & 0.2843***    & 0.2269***  & 0.0806***  & 0.0290         & 0.0298           \\
                                       & (0.0240)     & (0.0231)   & (0.0250)   & (0.0264)       & (0.0264)         \\
topic\_switching\_rate                 & 0.0996***    & 0.0951***  & 0.1424***  & 0.0390         & 0.0388           \\
                                       & (0.0301)     & (0.0287)   & (0.0282)   & (0.0307)       & (0.0307)         \\
trajectory\_curvature                  & -0.0564**    & -0.0394*   & -0.0192    & -0.0027        & -0.0030          \\
                                       & (0.0243)     & (0.0232)   & (0.0229)   & (0.0228)       & (0.0228)         \\
Adj. R²                                & 0.084        & 0.173      & 0.219      & 0.234          & —                \\
Conditional R²                         & —            & —          & —          & —              & 0.325            \\
Marginal R²                            & —            & —          & —          & —              & 0.208            \\
N                                      & 3574         & 3574       & 3574       & 3574           & 3574             \\
R²                                     & 0.086        & 0.176      & 0.223      & 0.238          & —                \\
\hline
\end{tabular}
\end{table}

\begin{table}
\caption{\textbf{Hierarchical regression models predicting human creativity from trajectory features.} 
Models 1-2: OLS regression with robust standard errors (HC3). 
Model 3: Mixed effects with task random intercepts (REML estimation). 
Outcome variable: standardized creativity rating (z-scored within human sample). 
All trajectory features standardized (z-scored). 
Conditional R² includes both fixed and random effects; Marginal R² includes fixed effects only. 
Reference category: Task = education inequality. 
$^{*}p<0.10$, $^{**}p<0.05$, $^{***}p<0.01$.}
\label{tab:human_regression_full}
\centering
\begin{tabular}{llll}
\hline
                                       & Model 1      & Model 2    & Model 3          \\
                                       & (Trajectory) & (+ Task)   & (Mixed Effects)  \\
\hline
C(question\_id)[T.employee\_attrition] &              & 0.0510     &                  \\
                                       &              & (0.1233)   &                  \\
C(question\_id)[T.plastic\_waste]      &              & -0.2754*   &                  \\
                                       &              & (0.1425)   &                  \\
C(question\_id)[T.singing\_shower]     &              & -0.4026*** &                  \\
                                       &              & (0.1338)   &                  \\
C(question\_id)[T.sorry\_pandemic]     &              & -0.0204    &                  \\
                                       &              & (0.1399)   &                  \\
C(question\_id)[T.supply\_chain]       &              & -0.4339*** &                  \\
                                       &              & (0.1336)   &                  \\
Group Var                              &              &            & 0.0437           \\
                                       &              &            & (0.0344)         \\
Intercept                              & -0.0000      & 0.1715*    & -0.0069          \\
                                       & (0.0382)     & (0.0910)   & (0.0881)         \\
R-squared                              & 0.1164       & 0.1527     &                  \\
R-squared Adj.                         & 0.1032       & 0.1329     &                  \\
convergence\_ratio                     & 0.0153       & 0.0127     & 0.0131           \\
                                       & (0.0385)     & (0.0385)   & (0.0383)         \\
global\_coherence                      & -0.5108***   & -0.4631*** & -0.4729***       \\
                                       & (0.1024)     & (0.1023)   & (0.1021)         \\
local\_coherence                       & 0.2726***    & 0.2329***  & 0.2412***        \\
                                       & (0.0688)     & (0.0692)   & (0.0691)         \\
max\_distance                          & 0.0098       & -0.0304    & -0.0224          \\
                                       & (0.0762)     & (0.0756)   & (0.0756)         \\
path\_length                           & -0.0511      & -0.0270    & -0.0322          \\
                                       & (0.0616)     & (0.0617)   & (0.0615)         \\
revisit\_score                         & -0.0926*     & -0.0576    & -0.0643          \\
                                       & (0.0476)     & (0.0479)   & (0.0478)         \\
semantic\_spread                       & 0.0270       & 0.0443     & 0.0410           \\
                                       & (0.0487)     & (0.0493)   & (0.0490)         \\
topic\_switching\_rate                 & 0.1385***    & 0.1095**   & 0.1152**         \\
                                       & (0.0471)     & (0.0469)   & (0.0469)         \\
trajectory\_curvature                  & 0.2234***    & 0.2065***  & 0.2101***        \\
                                       & (0.0514)     & (0.0511)   & (0.0510)         \\
Adj. R²                                & 0.103        & 0.133      & —                \\
Conditional R²                         & —            & —          & 0.170            \\
Marginal R²                            & —            & —          & 0.134            \\
N                                      & 308          & 308        & 308              \\
R²                                     & 0.116        & 0.153      & —                \\
\hline
\end{tabular}
\end{table}

\begin{table}[h]
\centering
\caption{\textbf{Model-specific variance explained by trajectory features.} 
Mixed effects models with task random intercepts. 
Trajectory features explain more variance for reasoning models (o3) than standard models (GPT-4.1).}
\label{tab:model_variance}
\begin{tabular}{lccc}
\hline
\textbf{Model} & \textbf{Trajectory Only} & \textbf{+ Task} & \textbf{Mixed Effects} \\
\hline
Mixed & 5.6\% & 12.4\% & 19.1\% \\
gpt-4.1 & 6.4\% & 17.8\% & 27.9\% \\
o3-high & 10.3\% & 30.7\% & \textbf{46.4\%} \\
o3-low & 7.8\% & 23.0\% & 36.1\% \\
\hline
\end{tabular}
\end{table}

\begin{table}[h]
\centering
\caption{\textbf{Mechanistic differences: LLM vs human trajectory feature coefficients.} 
Coefficients from OLS models with robust standard errors (Model 2: Trajectory + Task) predicting standardized creativity. 
Both models include trajectory features + task fixed effects for fair comparison. 
All features and outcomes standardized (z-scored within sample). 
LLM creativity is driven by semantic deviation magnitude (low global coherence, high semantic spread) with penalties for inefficient wandering (negative path length, max distance). Human creativity is driven by exploration structure (high local coherence, high trajectory curvature). 
$^{*}p<0.05$, $^{**}p<0.01$, $^{***}p<0.001$.}
\label{tab:llm_human_coefficients}
\begin{tabular}{lcccccccc}
\hline
\textbf{Feature} & \textbf{LLM $\beta$} & \textbf{SE} & \textbf{Sig} & \textbf{Human $\beta$} & \textbf{SE} & \textbf{Sig} & \textbf{Diff} \\
\hline
Local Coherence & -0.0527 & (0.0451) &  & 0.2329 & (0.0692) & *** & -0.2856 \\
Global Coherence & -0.4001 & (0.0399) & *** & -0.4631 & (0.1023) & *** & +0.0630 \\
Path Length & -0.1216 & (0.0207) & *** & -0.0270 & (0.0617) &  & -0.0946 \\
Convergence Ratio & -0.0226 & (0.0173) &  & 0.0127 & (0.0385) &  & -0.0353 \\
Max Distance & -0.2758 & (0.0278) & *** & -0.0304 & (0.0756) &  & -0.2454 \\
Trajectory Curvature & -0.0394 & (0.0232) &  & 0.2065 & (0.0511) & *** & -0.2458 \\
Topic Switching Rate & 0.0951 & (0.0287) & *** & 0.1095 & (0.0469) & * & -0.0145 \\
Revisit Score & 0.0086 & (0.0196) &  & -0.0576 & (0.0479) &  & +0.0662 \\
Semantic Spread & 0.2269 & (0.0231) & *** & 0.0443 & (0.0493) &  & +0.1826 \\
\hline
\end{tabular}
\end{table}

\begin{table}[h]
\centering
\caption{\textbf{Manipulability of trajectory features through experimental design.} 
Hierarchical OLS regression models with robust standard errors (HC3) showing variance explained (R²) in each trajectory feature by experimental design choices. 
Model 1: Task fixed effects only (baseline). 
Model 2: Task + Discussion Structure. 
Model 3: Task + Model Type. 
Model 4: Task + Discussion + Model (additive). 
Model 5: Task + Discussion × Model (with interaction). 
N=3,574 LLM conversations.}
\label{tab:trajectory_manipulability_detailed}
\begin{tabular}{lccccc}
\hline
\textbf{Trajectory Feature} & \textbf{Task} & \textbf{+Discussion} & \textbf{+Model} & \textbf{Additive} & \textbf{Interaction} \\
\hline
Global Coherence               &  5.9\% & 27.0\% & 28.8\% & 47.5\% & 52.6\% \\
Max Distance                   &  6.4\% & 47.0\% &  9.1\% & 48.9\% & 51.7\% \\
Local Coherence                &  0.5\% & 11.9\% & 31.9\% & 41.5\% & 43.4\% \\
Semantic Spread                &  0.4\% &  9.3\% &  2.7\% & 11.2\% & 24.8\% \\
Topic Switching Rate           &  0.3\% &  6.9\% & 13.1\% & 20.2\% & 24.0\% \\
Path Length                    &  0.1\% &  2.1\% &  9.0\% & 10.8\% & 14.0\% \\
Revisit Score                  &  1.4\% &  7.0\% &  5.3\% & 11.2\% & 12.9\% \\
Trajectory Curvature           &  0.4\% &  2.5\% &  5.4\% &  7.3\% &  9.6\% \\
Convergence Ratio              &  0.5\% &  5.7\% &  1.0\% &  6.2\% &  7.9\% \\
\hline
\textbf{Mean across features} &  1.8\% & 13.3\% & 11.8\% & 22.8\% & 26.8\% \\
\hline
\end{tabular}
\end{table}

\begin{table}[h]
\centering
\caption{\textbf{Interaction effects in trajectory feature manipulability.} 
Decomposition of variance explained into additive effects (discussion + model independently) and interaction effects (discussion × model). 
Features with high interaction percentages indicate that optimal discussion structures vary substantially by model type.}
\label{tab:trajectory_interaction_effects}
\begin{tabular}{lcccc}
\hline
\textbf{Trajectory Feature} & \textbf{Additive R²} & \textbf{Interaction R²} & \textbf{Full R²} & \textbf{\% from Interaction} \\
\hline
Semantic Spread                & 11.2\% & 13.6\% & 24.8\% &   55\% \\
Global Coherence               & 47.5\% &  5.1\% & 52.6\% &   10\% \\
Topic Switching Rate           & 20.2\% &  3.7\% & 24.0\% &   16\% \\
Path Length                    & 10.8\% &  3.3\% & 14.0\% &   23\% \\
Max Distance                   & 48.9\% &  2.8\% & 51.7\% &    5\% \\
Trajectory Curvature           &  7.3\% &  2.2\% &  9.6\% &   23\% \\
Local Coherence                & 41.5\% &  1.9\% & 43.4\% &    4\% \\
Revisit Score                  & 11.2\% &  1.7\% & 12.9\% &   14\% \\
Convergence Ratio              &  6.2\% &  1.7\% &  7.9\% &   22\% \\
\hline
\end{tabular}
\end{table}

\begin{table}[h]
\centering
\caption{\textbf{Robustness check: Alternative embedding model (Qwen3-Embedding-4B vs Qwen3-Embedding-0.6B).} 
Standardized coefficients from OLS models predicting creativity from trajectory features with task fixed effects.
Results are robust to embedding model choice, with consistent coefficient signs and significance patterns.
$^{*}p<0.05$, $^{**}p<0.01$, $^{***}p<0.001$.}
\label{tab:robustness_embedding}
\begin{tabular}{lcccc}
\hline
 & \multicolumn{2}{c}{\textbf{LLM}} & \multicolumn{2}{c}{\textbf{Human}} \\
\textbf{Feature} & 0.6B & 4B & 0.6B & 4B \\
\hline
Global Coherence & -0.400*** & -0.211*** & -0.463** & -0.495** \\
Semantic Spread & 0.227*** & 0.131*** & 0.044 & 0.006 \\
Path Length & -0.122*** & -0.114*** & -0.027 & 0.053 \\
Local Coherence & -0.053 & -0.077 & 0.233* & 0.296* \\
Trajectory Curvature & -0.039 & -0.034 & 0.206** & 0.174* \\
Max Distance & -0.276*** & -0.182*** & -0.030 & -0.113 \\
Topic Switching Rate & 0.095*** & 0.089** & 0.110 & 0.076 \\
Convergence Ratio & -0.023 & -0.032 & 0.013 & 0.042 \\
Revisit Score & 0.009 & 0.057** & -0.058 & -0.012 \\
\hline
$R^2$ & 17.6\% & 14.7\% & 15.3\% & 13.7\% \\
\hline
\end{tabular}
\end{table}

\begin{table}[p]
\centering
\caption{\textbf{Side-by-side conversation comparison illustrating semantic exploration patterns.} 
Each row shows the first sentence of each turn. 
High-creativity conversation (left) exhibits dramatic conceptual leaps across semantic domains (global coherence = -0.91), 
while low-creativity conversation (right) iteratively refines a single core concept (global coherence = 2.20). 
Both conversations use the same discussion structures (instructed) and model (GPT-4.1) for the same task (plastic waste).}
\label{tab:conversation_comparison}
\footnotesize
\begin{tabular}{cp{5.5cm}p{5.5cm}}
\toprule
\textbf{Turn} & \textbf{High Creativity (0.551)} & \textbf{Low Creativity (0.090)} \\
\midrule
1 & Modify: Create an integrated community-based micro-recycling hub network combined with a digital platform in apartment complexes and office buildings. & Modify: Create a “Plastic Waste-to-Art Currency” program: Establish local collection hubs where people exchange plastic waste for digital credits, which are redeemable for community goods, services, or access to public events. \\
2 & Modify: Transform the community-based micro-recycling hub network into an “Eco-Creation Lab” system, where each hub is not only for recycling plastic but also serves as a public maker-space and eco-innovation incubator. & Modify: Launch a “Plastic Blockchain Upcycling Ecosystem”: Establish decentralized plastic collection centers where individuals deposit plastic waste and, via AI-powered scanners, instantly receive blockchain-verified digital tokens. \\
3 & Modify: Evolve the “Eco-Creation Lab” system into an “Urban Plastic Circularity Accelerator” that integrates micro-recycling hubs with city-wide plastic tracking, AI-driven needs analysis, and an urban materials exchange marketplace. & Modify: Develop “Plastic Memory” Smart Packaging – All consumer plastics are embedded with ultra-thin, flexible NFC (Near Field Communication) chips that store the plastic’s entire lifecycle, production details, and recycling instructions. \\
4 & Modify: Transform the “Urban Plastic Circularity Accelerator” into an interactive city-wide “Plastic Hackathon Network. & Modify: Launch the “Plastic Genome Project”—a global initiative where every plastic item is manufactured with a microscopic, biodegradable DNA-based tag encoding its origin, composition, and recycling pathway. \\
5 & Modify: Evolve the “Plastic Hackathon Network” into a gamified, transmedia “Plastic Quest”—a citywide, augmented reality (AR)-powered challenge blending physical and digital worlds. & Agree: No changes needed. \\
6 & Modify: Transform “Plastic Quest” into a continuous, city-scale “Eco-Reality Game Show” that makes plastic reduction a collaborative, televised spectacle. & Modify: Establish “Plastic Zero Zones”—specialized, hyperlocal micro-factories powered by renewable energy, where plastics collected from the neighborhood are transformed onsite into eco-building materials, modular repair components, or public urban infrastructure using advanced low-emission processes. \\
7 & Modify: Amplify the “Eco-Reality Game Show” by transforming it into a globally interconnected, time-bounded “World Plastic Championship. & Modify: Create “Living Plastic Ecosystems” — Deploy modular, floating community gardens in urban waterways that actively consume and convert plastic waste using engineered plant-microbe symbiosis. \\
8 & Modify: Rather than a yearly “World Plastic Championship” that relies on city teams, transform the event into a “Global Plastic Innovation Relay”—a rapid, time-limited challenge where breakthrough plastic reduction ideas (whether tech, design, policy, or social) are crowd-sourced from every continent, then passed virtually from city to city like an Olympic torch. & Agree: No changes needed. \\
9 & Agree: No changes needed. & Agree: No changes needed. \\
10 & Agree: No changes needed. & Agree: No changes needed. \\
11 & Agree: No changes needed. & Agree: No changes needed. \\
12 & — & Agree: No changes needed. \\
13 & — & Agree: No changes needed. \\
\bottomrule
\end{tabular}
\end{table}

\begin{table}[h]
\centering
\caption{\textbf{Robustness check: Alternative creativity operationalization (Additive vs Multiplicative).} 
Standardized coefficients from OLS models predicting creativity from trajectory features with task fixed effects.
Multiplicative: Novelty $\times$ Usefulness. Additive: Novelty + Usefulness.
$^{*}p<0.05$, $^{**}p<0.01$, $^{***}p<0.001$.}
\label{tab:robustness_additive}
\begin{tabular}{lcccc}
\hline
 & \multicolumn{2}{c}{\textbf{LLM}} & \multicolumn{2}{c}{\textbf{Human}} \\
\textbf{Feature} & Mult & Add & Mult & Add \\
\hline
Global Coherence & -0.400*** & -0.434*** & -0.463** & -0.407** \\
Semantic Spread & 0.227*** & 0.219*** & 0.044 & 0.005 \\
Path Length & -0.122*** & -0.117*** & -0.027 & -0.049 \\
Local Coherence & -0.053 & -0.005 & 0.233* & 0.272** \\
Trajectory Curvature & -0.039 & -0.025 & 0.206** & 0.206** \\
Max Distance & -0.276*** & -0.291*** & -0.030 & 0.045 \\
Topic Switching Rate & 0.095*** & 0.091** & 0.110 & 0.085 \\
Convergence Ratio & -0.023 & -0.011 & 0.013 & 0.025 \\
Revisit Score & 0.009 & 0.038 & -0.058 & -0.043 \\
\hline
$R^2$ & 17.6\% & 19.0\% & 15.3\% & 15.0\% \\
\hline
\end{tabular}
\end{table}

\begin{table}[h]
\centering
\caption{\textbf{Robustness Check: Discussion structure effects survive token count control.} 
OLS regression predicting creativity from discussion structure, controlling for model type, task, and total token count.
$^{*}p<0.05$, $^{**}p<0.01$, $^{***}p<0.001$. N=4,120 LLM conversations with discussion.}
\label{tab:robustness_tokens}
\begin{tabular}{lc}
\hline
\textbf{Predictor} & \textbf{$\beta$} \\
\hline
Instructed (vs. progressive) & 0.002 \\
Iterative (vs. progressive) & 0.016** \\
Open (vs. progressive) & $-$0.030*** \\
Token count & $\approx$0 (n.s.) \\
\hline
$R^2$ & 24.1\% \\
\hline
\end{tabular}
\end{table}

\end{appendices}

\end{document}